\documentclass{article}

\usepackage{microtype}
\usepackage{graphicx}
\usepackage{subfigure}
\usepackage{booktabs} 

\usepackage{amssymb}
\usepackage{amstext}
\usepackage{mathtools}
\usepackage[normalem]{ulem}
\usepackage{caption}

\usepackage{hyperref}

\usepackage[accepted]{icml2019}

\icmltitlerunning{Extrapolating Beyond Suboptimal Demonstrations via Inverse Reinforcement Learning from Observations}

\begin{document}

\twocolumn[
\icmltitle{Extrapolating Beyond Suboptimal Demonstrations via \\ Inverse Reinforcement Learning from Observations}



\icmlsetsymbol{equal}{*}

\begin{icmlauthorlist}
\icmlauthor{Daniel S. Brown}{equal,ut}
\icmlauthor{Wonjoon Goo}{equal,ut}
\icmlauthor{Prabhat Nagarajan}{pn}
\icmlauthor{Scott Niekum}{ut}
\end{icmlauthorlist}

\icmlaffiliation{ut}{Department of Computer Science, University of Texas at Austin, USA}
\icmlaffiliation{pn}{Preferred Networks, Japan}

\icmlcorrespondingauthor{Daniel S. Brown}{dsbrown@cs.utexas.edu}
\icmlcorrespondingauthor{Wonjoon Goo}{wonjoon@cs.utexas.edu}

\icmlkeywords{inverse reinforcement learning, IRL, ICML}

\vskip 0.3in]



\printAffiliationsAndNotice{\icmlEqualContribution} 

\begin{abstract}
A critical flaw of existing inverse reinforcement learning (IRL) methods is their inability to significantly outperform the demonstrator. 
This is because IRL typically seeks a reward function that makes the demonstrator appear near-optimal, rather than inferring the underlying intentions of the demonstrator that may have been poorly executed in practice.
In this paper, we introduce a novel reward-learning-from-observation algorithm, Trajectory-ranked Reward EXtrapolation (T-REX), that extrapolates beyond a set of (approximately) ranked demonstrations in order to infer high-quality reward functions from a set of potentially poor demonstrations. When combined with deep reinforcement learning, T-REX outperforms state-of-the-art imitation learning and IRL methods on multiple Atari and MuJoCo benchmark tasks and achieves performance that is often more than twice the performance of the best demonstration. 
We also demonstrate that T-REX is robust to ranking noise and can accurately extrapolate intention by simply watching a learner noisily improve at a task over time.

\end{abstract}

\section{Introduction}

\begin{figure}
    \centering
    \includegraphics[width=1\linewidth]{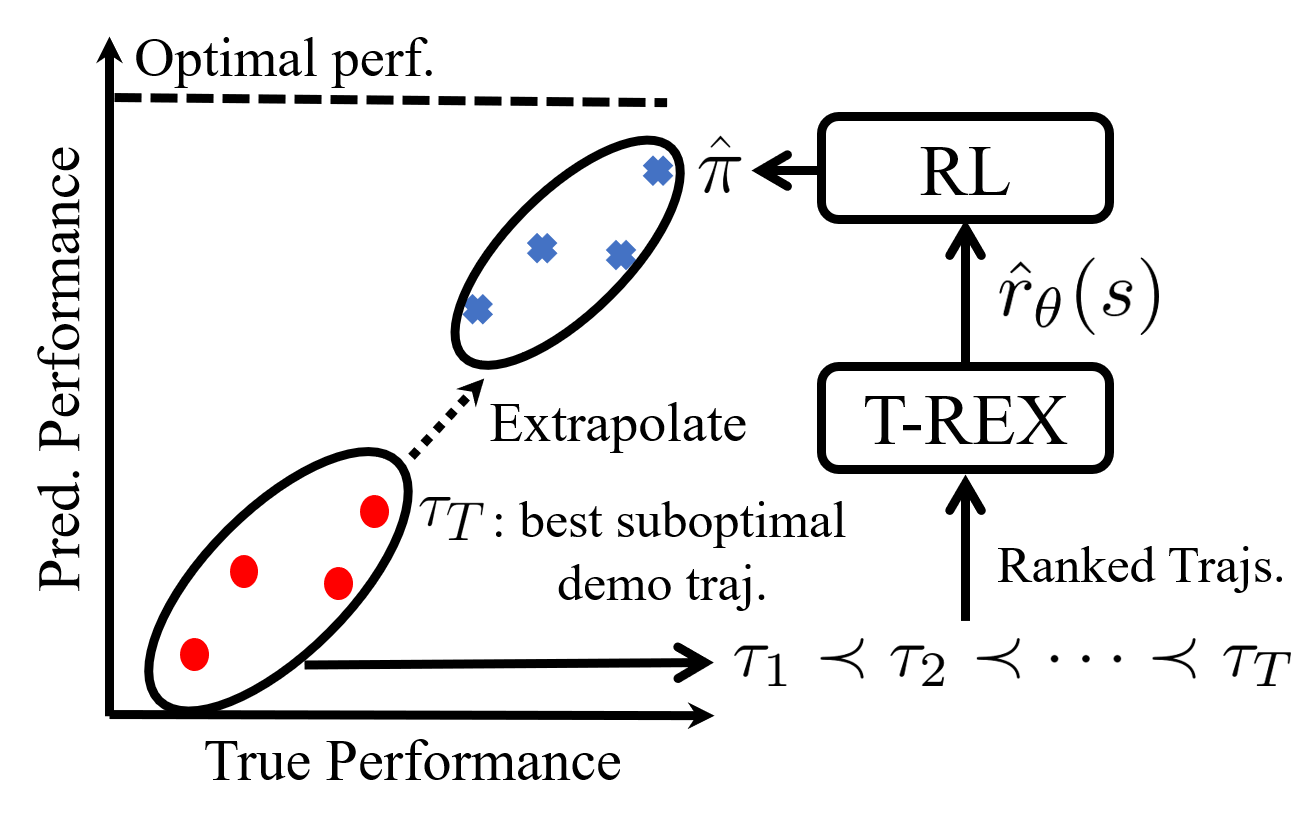}
    \caption{T-REX takes a sequence of ranked demonstrations and learns a reward function from these rankings that allows policy improvement over the demonstrator via reinforcement learning.}
    \label{fig:architecture}
    \vspace{-0.3in}
\end{figure}

Due to advantages such as computational speed, precise manipulation, and exact timing, computers and robots are often superior to humans at performing tasks with well-defined goals and objectives. However, it can be difficult, even for experts, to design reward functions and objectives that lead to desired behaviors when designing autonomous agents \cite{ng1999policy,amodei2016concrete}. When goals or rewards are difficult for a human to specify, inverse reinforcement learning (IRL) \cite{abbeel2004apprenticeship} techniques can be applied to infer the intrinsic reward function of a user from demonstrations. 
Unfortunately, high-quality demonstrations are difficult to provide for many tasks---for instance, consider 
a non-expert user attempting to give kinesthetic demonstrations of a household chore to a robot. Even for relative experts, tasks such as high-frequency stock trading or playing complex video games can be difficult to perform optimally.

If a demonstrator is suboptimal, but their intentions can be ascertained, then a learning agent ought to be able to exceed the demonstrator's performance in principle. However, existing IRL algorithms fail to do this, typically searching for a reward function that makes the demonstrations appear near-optimal \cite{ramachandran2007bayesian,ziebart2008maximum,finn2016guided,henderson2018optiongan}. Thus, when the demonstrator is suboptimal, IRL results in suboptimal behavior as well.  Imitation learning approaches \cite{Argall2009} that mimic behavior directly without reward inference, such as behavioral cloning \cite{torabi2018behavioral}, also suffer from the same shortcoming. 

To overcome this critical flaw in current imitation learning methods, we propose a novel IRL algorithm, Trajectory-ranked Reward EXtrapolation (T-REX)\footnote{Code available at \url{https://github.com/hiwonjoon/ICML2019-TREX}} that utilizes ranked demonstrations to extrapolate a user's underlying intent beyond the best demonstration, even when all demonstrations are highly suboptimal. This, in turn, enables a reinforcement learning agent to exceed the performance of the demonstrator by learning to optimize this extrapolated reward function. Specifically, we use ranked demonstrations to learn a state-based reward function that assigns greater total return to higher-ranked trajectories. Thus, while standard inverse reinforcement learning approaches seek a reward function that \textit{justifies} the demonstrations, we instead seek a reward function that \textit{explains} the ranking over demonstrations, allowing for potentially better-than-demonstrator performance.

  Utilizing ranking in this way has several advantages.  First, rather than imitating suboptimal demonstrations, it allows us to identify features that are correlated with rankings, in a manner that can be extrapolated beyond the demonstrations.  
  Although the learned reward function could potentially overfit to the provided rankings, we demonstrate empirically that it extrapolates well, successfully predicting returns of trajectories that are significantly better than any observed demonstration, likely due to the powerful regularizing effect of having many pairwise ranking constraints between trajectories. For example, the degenerate all-zero reward function (the agent always receives a reward of 0) makes any given set of demonstrations appear optimal. However, such a reward function is eliminated from consideration by any pair of (non-equally) ranked demonstrations. Second, when learning features directly from high-dimensional data, this regularizing effect can also help to prevent overfitting to the small fraction of state space visited by the demonstrator.
  By utilizing a set of suboptimal, but ranked demonstrations, we provide the neural network with diverse data from multiple areas of the state space, allowing an agent to better learn both what to do and what not to do in a variety of situations.

We evaluate T-REX on a variety of standard Atari and MuJoCo benchmark tasks.  Our experiments show that T-REX can extrapolate well, achieving performance that is often more than twice as high as the best-performing demonstration, as well as outperforming state-of-the-art imitation learning algorithms. 
We also show that T-REX performs well even in the presence of significant ranking noise, and provide results showing that T-REX can learn good policies simply by observing a novice demonstrator that noisily improves over time.

\section{Related Work}
The goal of our work is to achieve improvements over a sub-optimal demonstrator in high-dimensional reinforcement learning tasks without requiring a hand-specified reward function or supervision during policy learning. While there is a large body of research on learning from demonstrations \cite{Argall2009,gao2012survey,osa2018algorithmic,arora2018survey}, most work assumes access to action labels, while we learn only from observations. Additionally, little work has addressed the problem of learning from ranked demonstrations, especially when they are significantly suboptimal. To the best of our knowledge, our work is the first to show better-than-demonstrator performance in high-dimensional tasks such as Atari, without requiring active human supervision or access to ground-truth rewards.

\subsection{Learning from demonstrations}
Early work on learning from demonstration focused on behavioral cloning \cite{pomerleau1991efficient}, in which the goal is to learn a policy that imitates the actions taken by the demonstrator; however, without substantial human feedback and correction, this method is known to have large generalization error \cite{ross2011reduction}. Recent deep learning approaches to imitation learning \cite{ho2016generative} have used Generative Adversarial Networks \cite{goodfellow2014generative} to model the distribution of actions taken by the demonstrator.

Rather than directly learn to mimic the demonstrator, inverse reinforcement learning (IRL) \cite{gao2012survey,arora2018survey} seeks to find a reward function that models the intention of the demonstrator, thereby allowing generalization to states that were unvisited during demonstration. Given such a reward function, reinforcement learning \cite{sutton1998introduction} techniques can be applied to learn an optimal policy.
Maximum entropy IRL seeks to find a reward function that makes the demonstrations appear near-optimal, while further disambiguating inference by also maximizing the entropy of the resulting policy \cite{ziebart2008maximum,boularias2011relative,wulfmeier2015maximum,finn2016guided}. While maximum entropy approaches are robust to limited and occasional suboptimality in the demonstrations, they still fundamentally seek a reward function that justifies the demonstrations, resulting in performance that is explicitly tied to the performance of the demonstrator.

\citet{syed2008game} proved that, given prior knowledge about which features contribute positively or negatively to the true reward, an apprenticeship policy can be found that is guaranteed to outperform the demonstrator. However, their approach requires hand-crafted, linear features, knowledge of the true signs of the rewards features, and also requires repeatedly solving a Markov decision process (MDP). Our proposed method uses deep learning and ranked demonstrations to automatically learn complex features that are positively and negatively correlated with performance, and is able to generate a policy that can outperform the demonstrator via the solution to a single RL problem.

Our work can be seen as a form of preference-based policy learning \cite{akrour2011preference} and  preference-based IRL (PBIRL) \cite{pbirl,irldialog} which both seek to optimize a policy based on preference rankings over demonstrations. However, existing approaches only consider reward functions that are linear in hand-crafted features and have not studied extrapolation capabilities. For a more complete overview survey of preference-based reinforcement learning, see the survey by \citet{preferencesurvey}. Other methods \cite{dmirl, scorebasedirl} have proposed the use of quantitatively scored trajectories as opposed to qualitative pairwise preferences over demonstrations. However, none of the aforementioned methods have been applied to the types of high-dimensional deep inverse reinforcement learning tasks considered in this paper.

\subsection{Learning from observation}
Recently there has been a shift towards learning from observations, in which the actions taken by the demonstrator are unknown. \citet{torabi2018behavioral} propose a state-of-the-art model-based approach to perform behavioral cloning from observation. \citet{sermanet2018time} and \citet{liu2018imitation} propose methods to learn directly from a large corpus of videos containing multiple view points of the same task. \citet{yu2018one} and \citet{goo2019one} propose meta-learning-from-observation approaches that can learn from a single demonstration, but require training on a wide variety of similar tasks. \citet{henderson2018optiongan} and \citet{torabi2018generative} extend Generative Adversarial Imitation Learning \cite{ho2016generative} to remove the need for action labels. However, inverse reinforcement learning methods based on Generative Adversarial Networks \cite{goodfellow2014generative} are notoriously difficult to train and have been shown to fail to scale to high-dimensional imitation learning tasks such as Atari \cite{irlvideogames}.

\subsection{Learning from suboptimal demonstrations}

Very little work has tried to learn good policies from highly suboptimal demonstrations. \citet{grollman2011donut} propose a method that learns from failed demonstrations where a human attempts, but is unable, to perform a task; however, demonstrations must be labeled as failures and manually clustered into two sets of demonstrations: those that overshoot and those that undershoot the goal. 
\citet{shiarlis2016inverse} demonstrate that if successful and failed demonstrations are labeled and the reward function is a linear combination of known features, then maximum entropy IRL can be used to optimize a policy to match the expected feature counts of successful demonstrations while not matching the feature counts of failed demonstrations. \citet{zheng2014robust} and \citet{choi2019robust} propose methods that are robust to small numbers of unlabeled suboptimal demonstrations, but require a majority of expert demonstrations in order to correctly identify which demonstrations are anomalous. 

In reinforcement learning, it is common to initialize a policy from suboptimal demonstrations and then improve this policy using the ground truth reward signal \cite{kober2009policy,taylor2011integrating,hester2017deep,gao2018reinforcement}. However, it is often still difficult to perform significantly better than the demonstrator \cite{hester2017deep} and designing reward functions for reinforcement learning can be extremely difficult for non-experts and can easily lead to unintended behaviors \cite{ng1999policy,amodei2016concrete}.

\subsection{Reward learning for video games}
Most deep learning-based methods for reward learning require access to demonstrator actions and do not scale to high-dimensional tasks such as video games (e.g. Atari) \cite{ho2016generative,finn2016guided,airl,qureshi2018adversarial}. \citet{irlvideogames} tested state-of-the-art IRL methods on the Atari domain and showed that they are unsuccessful, even with near-optimal demonstrations and extensive parameter tuning. 

Our work builds on the work of \citet{christiano2017deep}, who proposed an algorithm that learns to play Atari games via pairwise preferences over trajectories that are actively collected during policy learning. However, this approach requires obtaining thousands of labels through constant human supervision during policy learning. In contrast, our method only requires an initial set of (approximately) ranked demonstrations as input and can learn a better-than-demonstrator policy without any supervision during policy learning. \citet{ibarz2018reward} combine deep Q-learning from demonstrations (DQfD) \cite{hester2017deep} and active preference learning \cite{christiano2017deep} to learn to play Atari games using both demonstrations and active queries. However, \citet{ibarz2018reward} require access to the demonstrator's actions in order to optimize an action-based, large-margin loss \cite{hester2017deep} and to optimize the state-action Q-value function using $(s,a,s')$-tuples from the demonstrations. Additionally, the large-margin loss encourages Q-values that make the demonstrator's actions better than alternative actions, resulting in performance that is often significantly worse than the demonstrator despite using thousands of active queries during policy learning.

\citet{imitationyoutube} use video demonstrations of experts to learn good policies for the Atari domains of Montezuma's Revenge, Pitfall, and Private Eye. Their method first learns a state-embedding and then selects a set of checkpoints from a demonstration. During policy learning, the agent is rewarded only when it reaches these checkpoints. This approach relies on high-performance demonstrations, which their method is unable to outperform. Furthermore, while \citet{imitationyoutube} do learn a reward function purely from observations, their method is inherently different from ours in that their learned reward function is designed to only imitate the demonstrations, rather than extrapolate beyond the capabilities of the demonstrator.

To the best of our knowledge, our work is the first to significantly outperform a demonstrator without using ground truth rewards or active preference queries. Furthermore, our approach does not require demonstrator actions and is able to learn a reward function that matches the demonstrator's intention without any environmental interactions---given rankings, reward learning becomes a binary classification problem and does not require access to an MDP.

\section{Problem Definition}
We model the environment as a Markov decision process (MDP) consisting of a set of states $\mathcal{S}$, actions $\mathcal{A}$, transition probabilities $P$, reward function $r: \mathcal{S} \rightarrow \mathbb{R}$, and discount factor $\gamma$ \cite{puterman2014markov}. A policy $\pi$ is a mapping from states to probabilities over actions, $\pi(a | s) \in [0,1]$. Given a policy and an MDP, the expected discounted return of the policy is given by $J(\pi) = \mathbf{E}[\sum_{t=0}^\infty \gamma^t r_t | \pi]$.

In this work we are concerned with the problem of inverse reinforcement learning from observation, where we do not have access to the reward function of the MDP nor the actions taken by the demonstrator.
Given a sequence of $m$ ranked trajectories $\tau_t$ for $t = 1,\ldots,m$, where $\tau_i \prec \tau_j$ if $i < j$, we wish to find a parameterized reward function $\hat{r}_\theta$ that approximates the true reward function $r$ that the demonstrator is attempting to optimize. Given $\hat{r}_\theta$, we then seek to optimize a policy $\hat{\pi}$ that can outperform the demonstrations.

We only assume access to a qualitative ranking over demonstrations. Thus, we only require the demonstrator to have an internal goal or intrinsic reward. The demonstrator can rank trajectories using any method, such as giving pairwise preferences over demonstrations or by rating each demonstration on a scale. Note that even if the relative scores of the demonstrations are used for ranking, it is still necessary to infer why some trajectories are better than others, which is what our proposed method does.

\section{Method}

We now describe Trajectory-ranked Reward EXtrapolation (T-REX), an algorithm for using ranked suboptimal demonstrations to extrapolate a user's underlying intent beyond the best demonstration. 
Given a sequence of $m$ demonstrations ranked from worst to best, $\tau_1, \ldots, \tau_m$, T-REX has two steps: (1) reward inference and (2) policy optimization. 

Given the ranked demonstrations, T-REX performs reward inference by approximating the reward at state $s$ using a neural network, $\hat{r}_\theta(s)$, such that $\sum_{s \in \tau_i} \hat{r}_\theta(s) < \sum_{s \in \tau_j} \hat{r}_\theta (s)$ when $\tau_i \prec \tau_j$.
The parameterized reward function $\hat{r}_\theta$ can be trained with ranked demonstrations using the generalized loss function:
\begin{equation}
    \mathcal{L}(\theta) = \mathbf{E}_{\tau_i,\tau_j \sim \Pi} \Big[ \xi \Big( \text{P} \big( \hat{J}_\theta(\tau_i) < \hat{J}_\theta(\tau_j) \big), \tau_i \prec \tau_j \Big) \Big],
\end{equation}
where $\Pi$ is a distribution over demonstrations, $\xi$ is a binary classification loss function, $\hat{J}$ is a (discounted) return defined by a parameterized reward function $\hat{r}_\theta$, and $\prec$ is an indication of the preference between the demonstrated trajectories.

We represent the probability $\text{P}$ as a softmax-normalized distribution and we instantiate $\xi$ using a cross entropy loss:
\begin{equation}
\text{P} \big( \hat{J}_\theta(\tau_i) < \hat{J}_\theta(\tau_j) \big) \approx \frac{\exp \displaystyle\sum_{s \in \tau_j} \hat{r}_\theta(s)}{\exp \displaystyle\sum_{s \in \tau_i} \hat{r}_\theta(s) + \exp \displaystyle\sum_{s \in \tau_j} \hat{r}_\theta(s)},
\end{equation}
\begin{equation}
 \mathcal{L}(\theta) = -\sum_{\tau_i \prec \tau_j} \log \frac{\exp \displaystyle\sum_{s \in \tau_j} \hat{r}_\theta(s)}{\exp \displaystyle\sum_{s \in \tau_i} \hat{r}_\theta(s) + \exp \displaystyle\sum_{s \in \tau_j} \hat{r}_\theta(s)}.
\end{equation}
This loss function trains a classifier that can predict whether one trajectory is preferable to another based on the predicted returns of each trajectory. This form of loss function follows from the classic Bradley-Terry and Luce-Shephard models of preferences \cite{bradley1952rank,luce2012individual} and has been shown to be effective for training neural networks from preferences \cite{christiano2017deep,ibarz2018reward}. 

To increase the number of training examples, T-REX trains on partial trajectory pairs rather than full trajectory pairs. 
This results in noisy preference labels that are only weakly supervised; however, using data augmentation to obtain pairwise preferences over many partial trajectories allows T-REX to learn expressive neural network reward functions from only a small number of ranked demonstrations. During training we randomly select pairs of trajectories, $\tau_{i}$ and $\tau_{j}$. We then randomly select partial trajectories $\tilde{\tau}_i$ and $\tilde{\tau}_j$ of length $L$. For each partial trajectory, we take each observation and perform a forward pass through the network $\hat{r}_\theta$ and sum the predicted rewards to compute the cumulative return. We then use the predicted cumulative returns as the logit values in the cross-entropy loss with the label corresponding to the higher ranked demonstration.

Given the learned reward function $\hat{r}_\theta(s)$, T-REX then seeks to optimize a policy $\hat{\pi}$ with better-than-demonstrator performance through reinforcement learning using $\hat{r}_\theta$.

\section{Experiments and Results}

\subsection{Mujoco}
We first evaluated our proposed method on three robotic locomotion tasks using the Mujoco simulator \cite{todorov2012mujoco} within OpenAI Gym \cite{brockman2016openai}, namely HalfCheetah, Hopper, and Ant. In all three tasks, the goal of the robot agent is to move forward as fast as possible without falling to the ground.

\subsubsection{Demonstrations}
To generate demonstrations, we trained a Proximal Policy Optimization (PPO) \cite{schulman2017proximal} agent with the ground-truth reward for 500 training steps (64,000 simulation steps) and checkpointed its policy after every 5 training steps. For each checkpoint, we generated a trajectory of length 1,000. This provides us with different demonstrations of varying quality which are then ranked based on the ground truth returns. To evaluate the effect of different levels of suboptimality, we divided the trajectories into different overlapping stages. We used 3 stages for HalfCheetah and Hopper. For HalfCheetah, we used the worst 9, 12, and 24 trajectories, respectively. For Hopper, we used the worst 9, 12, and 18 trajectories. For Ant, we used two stages consisting of the worst 12 and 40 trajectories. We used the PPO implementation from OpenAI Baselines \cite{baselines} with the given default hyperparameters.

\subsubsection{Experimental Setup}
We trained the reward network using 5,000 random pairs of partial trajectories of length 50, with preference labels based on the trajectory rankings, not the ground-truth returns. To prevent overfitting, we represented the reward function using an ensemble of five deep neural networks, trained separately with different random pairs. Each network has 3 fully connected layers of 256 units with ReLU nonlinearities. We train the reward network using the Adam optimizer \cite{kingma2014adam} with a learning rate of 1e-4 and a minibatch size of 64 for 10,000 timesteps.

To evaluate the quality of our learned reward, we then trained a policy to maximize the inferred reward function via PPO. The outputs of each the five reward networks in our ensemble, $\hat{r}(s)$, are normalized by their standard deviation to compensate for any scale differences amongst the models. The reinforcement learning agent receives the average of the ensemble as the reward, plus the control penalty used in OpenAI Gym \cite{brockman2016openai}. This control penalty represents a standard safety prior over reward functions for robotics tasks, namely to minimize joint torques. We found that optimizing a policy based solely on this control penalty does not lead to forward locomotion, thus learning a reward function from demonstrations is still necessary.

\subsubsection{Results}

\paragraph{Learned Policy Performance}

We measured the performance of the policy learned by T-REX by measuring the forward distance traveled. We also compared against Behavior Cloning from Observations (BCO) \cite{torabi2018behavioral}, a state-of-the-art learning-from-observation method, and Generative Adversarial Imitation Learning (GAIL) \cite{ho2016generative}, a state-of-the-art inverse reinforcement learning algorithm. BCO trains a policy via supervised learning, and has been shown to be competitive with state-of-the-art IRL \cite{ho2016generative} on MuJoCo tasks without requiring action labels, making it one of the strongest baselines when learning from observations. We trained BCO using only the best demonstration among the available suboptimal demonstrations. We trained GAIL with all of the demonstrations. GAIL uses demonstrator actions, while T-REX and BCO do not.     

We compared against three different levels of suboptimality (Stage 1, 2, and 3), corresponding to increasingly better demonstrations. The results are shown in Figure~\ref{fig:mujoco_rl_results} (see the appendix for full details). 
The policies learned by T-REX perform significantly better than the provided suboptimal trajectories in all the stages of HalfCheetah and Hopper. This provides evidence that T-REX can discover reward functions that extrapolate beyond the performance of the demonstrator. T-REX also outperforms BCO and GAIL on all tasks and stages except for Stage 2 for Hopper and Ant. BCO and GAIL usually fail to perform better than the average demonstration performance because they explicitly seek to imitate the demonstrator rather than infer the demonstrator's intention.

\begin{figure}[t]
    \centering
    \includegraphics[width=1.0\linewidth]{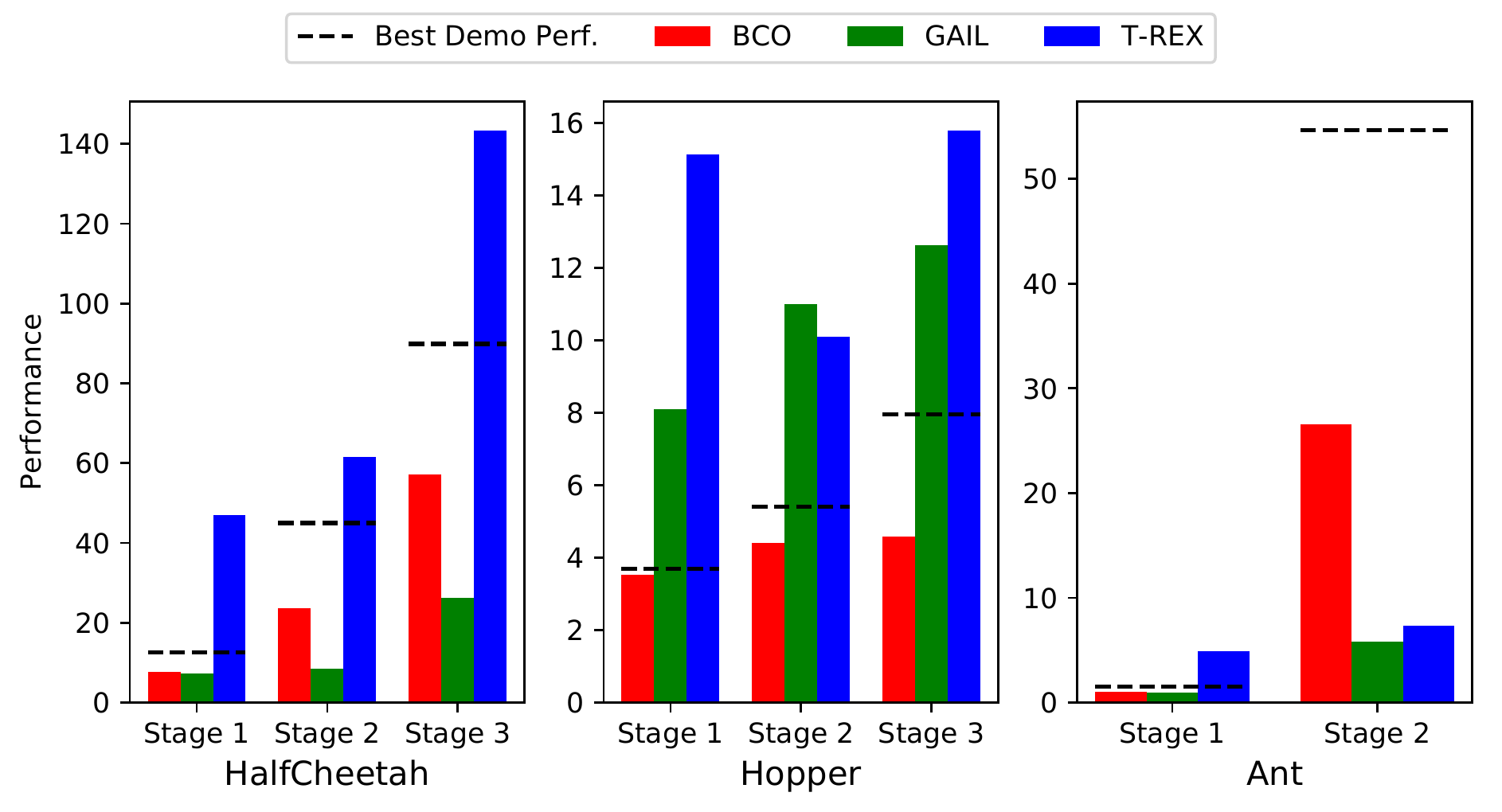}
    \caption{Imitation learning performance for three robotic locomotion tasks when given suboptimal demonstrations. Performance is measured as the total distance traveled, as measured by the final x-position of the robot's body. For each stage and task, the best performance given suboptimal demonstrations is shown for T-REX (ours), BCO \cite{torabi2018behavioral}, and GAIL \cite{ho2016generative}. The dashed line shows the performance of the best demonstration.}
    \label{fig:mujoco_rl_results}
\end{figure}

\begin{figure*}
    \centering
    \subfigure[HalfCheetah]{
        \includegraphics[width=0.26\linewidth]{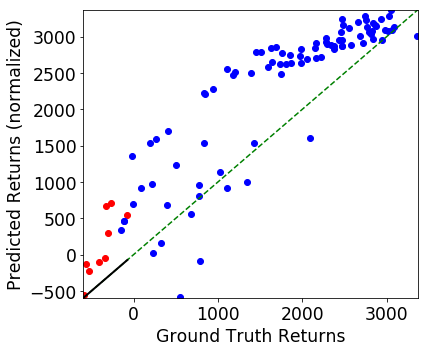}
        \label{subfig:halfcheetah_reward}
    }
    \subfigure[Hopper]{
        \includegraphics[width=0.26\linewidth]{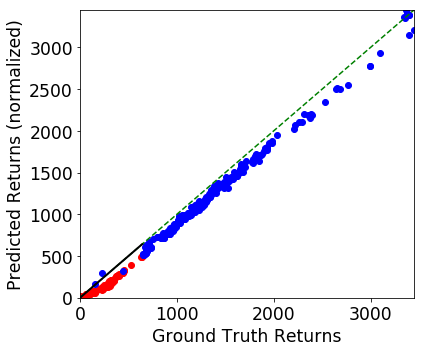}
        \label{subfig:hopper_reward}
    }
    \subfigure[Ant]{
        \includegraphics[width=0.26\linewidth]{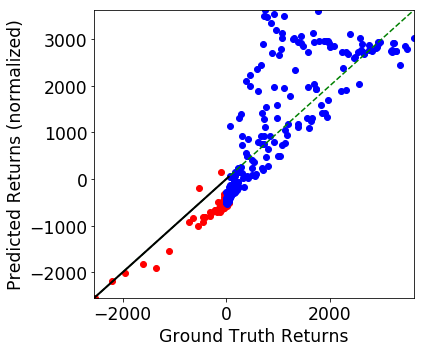}
        \label{subfig:ant_reward}
    }
    \caption{Extrapolation plots for T-REX on MuJoCo Stage 1 demonstrations. Red points correspond to demonstrations and blue points correspond to trajectories not given as demonstrations. The solid line represents the performance range of the demonstrator, and the dashed line represents extrapolation beyond the demonstrator's performance. The x-axis is the ground-truth return and the y-axis is the predicted return from our learned reward function.  Predicted returns are normalized to have the same scale as the ground-truth returns.}
    \label{fig:mujoco_reward_comparison}
\end{figure*}

\textbf{Reward Extrapolation} We next investigated the ability of T-REX to accurately extrapolate beyond the demonstrator. To do so, we compared ground-truth return and T-REX-inferred return across trajectories from a range of performance qualities, including trajectories much better than the best demonstration given to T-REX. The extrapolation of the reward function learned by T-REX is shown in  Figure~\ref{fig:mujoco_reward_comparison}. The plots in Figure~\ref{fig:mujoco_reward_comparison} give insight into the performance of T-REX. When T-REX learns a reward function that has a strong positive correlation with the ground-truth reward function, then it is able to surpass the performance of the suboptimal demonstrations. However, in Ant the correlation is not as strong, resulting in worse-than-demonstrator performance in Stage 2.

\subsection{Atari}

\subsubsection{Demonstrations}
We next evaluated T-REX on eight Atari games shown in Table~\ref{tab:atariLearningFromNovice}. 
To obtain a variety of suboptimal demonstrations, we generated 12 full-episode trajectories using PPO policies checkpointed every 50 training updates for all games except for Seaquest and Enduro. For Seaquest, we used every 5th training update due to the ability of PPO to quickly find a good policy. For Enduro, we used every 50th training update starting from step 3,100 since PPO obtained 0 return until after 3,000 steps. We used the OpenAI Baselines implementation of PPO with the default hyperparameters.

\subsubsection{Experimental Setup}
We used an architecture for reward learning similar to the one proposed in \cite{ibarz2018reward}, with four convolutional layers with sizes 7x7, 5x5, 3x3, and 3x3, with strides 3, 2, 1, and 1. Each convolutional layer used 16 filters and LeakyReLU non-linearities. We then used a fully connected layer with 64 hidden units and a single scalar output. We fed in stacks of 4 frames with pixel values normalized between 0 and 1 and masked the game score and number of lives.

For all games except Enduro, we subsampled 6,000 trajectory pairs between 50 and 100 observations long. We optimized the reward functions using Adam with a learning rate of 5e-5 for 30,000 steps. Given two full trajectories $\tau_i$ and $\tau_j$ such that $\tau_i \prec \tau_j$, we first randomly sample a subtrajectory from $\tau_i$. Let $t_i$ be the starting timestep for this subtrajectory. We then sample an equal length subtrajectory from $\tau_j$ such that $t_i \leq t_j$, where $t_j$ is the starting time step of the subtrajectory from $\tau_j$. We found that this resulted in better performance than comparing randomly chosen subtrajectories, likely due to the fact that (1) it eliminates pairings that compare a later part of a worse trajectory with an earlier part of a better trajectory and (2) it encourages reward functions that are monotonically increasing as progress is made in the game. For Enduro, training on short partial trajectories was not sufficient to score any points and instead we used 2,000 pairs of down-sampled full trajectories (see appendix for details). 

We optimized a policy by training a PPO agent on the learned reward function. To reduce reward scaling issues, we normalized predicted rewards by feeding the output of $\hat{r}_\theta(s)$ through a sigmoid function before passing it to PPO. We trained PPO on the learned reward function for 50 million frames to obtain our final policy. We also compare against Behavioral Cloning from Observation (BCO) \cite{torabi2018behavioral} and the state-of-the-art Generative Adversarial Imitation Learning (GAIL) \cite{ho2016generative}. Note that we give action labels to GAIL, but not to BCO or T-REX. We tuned the hyperparameters for GAIL to maximize performance when using expert demonstrations on Breakout and Pong. We gave the same demonstrations to both BCO and T-REX; however, we found that GAIL was very sensitive to poor demonstrations so we trained GAIL on 10 demonstrations using the policy checkpoint that generated the best demonstration given to T-REX.



\subsubsection{Results}
\paragraph{Learned Policy Performance}

\begin{table*}[ht!]
\caption{Comparison of T-REX with a state-of-the-art behavioral cloning algorithm (BCO) \cite{torabi2018behavioral} and state-of-the-art IRL algorithm (GAIL) \cite{ho2016generative}. Performance is evaluated on the ground-truth reward. T-REX achieves better-than-demonstrator performance on 7 out of 8 games and surpasses the BCO and GAIL baselines on 7 out of 8 games. Results are the best average performance over 3 random seeds with 30 trials per seed.}
\label{tab:atariLearningFromNovice}
\vskip 0.15in
\begin{center}
\begin{small}
\begin{tabular}{c|cc|ccccccc}
\toprule
 & \multicolumn{2}{c|}{Ranked Demonstrations} & \multicolumn{3}{c}{LfD Algorithm Performance} \\
 \midrule
Game &  Best & Average & T-REX & BCO & GAIL\\ 
\midrule 
Beam Rider &	1,332 &	686.0 &	\textbf{3,335.7} & 568  &  355.5 \\
Breakout &	32 &	14.5 &	\textbf{221.3} & 13  &  0.28\\
Enduro &	84 &	39.8 &		\textbf{586.8} & 8  &  0.28 \\
Hero &	\textbf{13,235} &	6,742.0 &	0 & 2,167 & 0 \\ 
Pong &	-6 &	-15.6 &	\textbf{-2.0} & -21 & -21 \\ 
Q*bert &	800 &	627 &	\textbf{32,345.8} & 150 & 0 \\ 
Seaquest &	600 &	373.3 &	\textbf{747.3} & 0  &  0\\ 
Space Invaders &	600 &	332.9 &	\textbf{1,032.5} & 88 &  370.2\\ 
\bottomrule
\end{tabular}
\end{small}
\end{center}
\vskip -0.1in
\end{table*}

The average performance of T-REX under the ground-truth reward function and the best and average performance of the demonstrator are shown in Table~\ref{tab:atariLearningFromNovice}.  
Table~\ref{tab:atariLearningFromNovice} shows that T-REX outperformed both BCO and GAIL in 7 out of 8 games. T-REX also outperformed the best demonstration in 7 out of 8 games. On four games (Beam Rider, Breakout, Enduro, and Q*bert) T-REX achieved score that is more than double the score of the best demonstration. In comparison, BCO performed worse than the average performance of the demonstrator in all games, and GAIL only performed better than the average demonstration on Space Invaders. Despite using better training data, GAIL was unable to learn good policies on any of the Atari tasks. These results are consistent with those of \citet{irlvideogames} that show that current GAN-based IRL methods do not perform well on Atari. In the appendix, we compare our results against prior work \cite{ibarz2018reward} that uses demonstrations plus active feedback during policy training to learn control policies for the Atari domain.

\begin{figure*}[ht!]
\centering
\subfigure[Beam Rider]{
\includegraphics[scale=0.24]{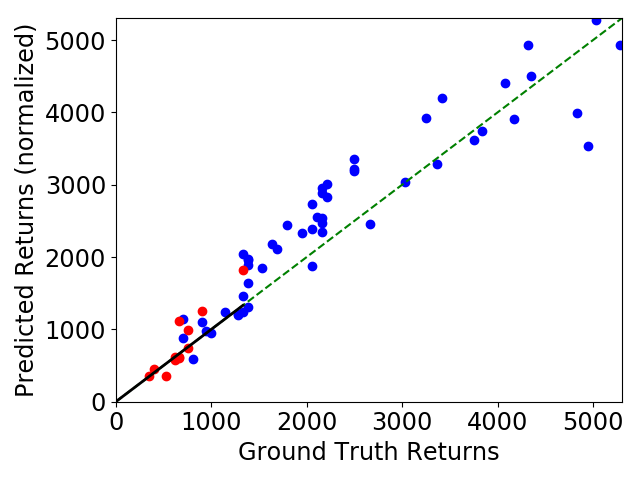}
\label{subfig:simpleNg}
}
\subfigure[Breakout]{
\includegraphics[scale=0.24]{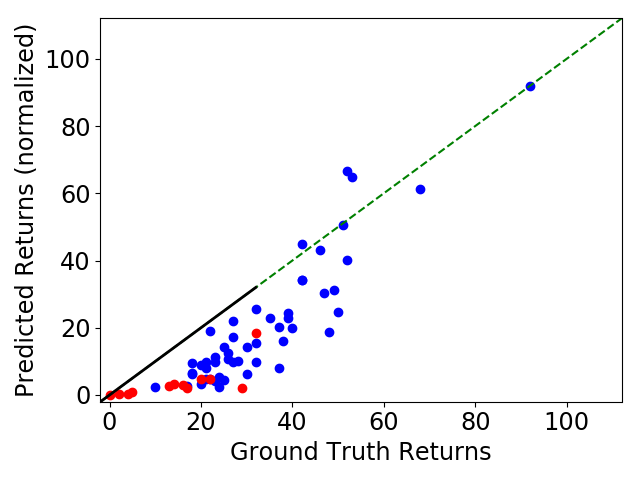}
\label{subfig:simplePolcicyFeasible}
}
\subfigure[Enduro]{
\includegraphics[scale=0.24]{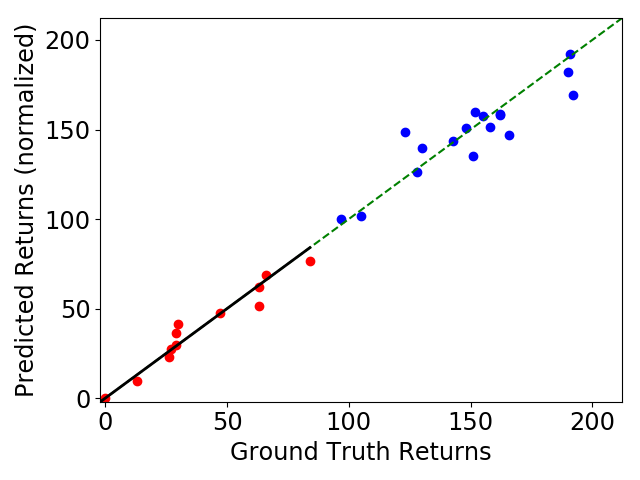}
\label{subfig:simpleCakmak}
}
\subfigure[Hero]{
\includegraphics[scale=0.24]{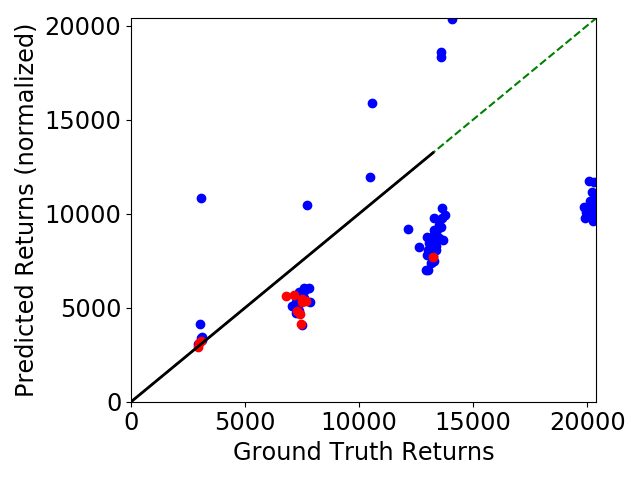}
\label{subfig:simpleCakmak2}
}
\subfigure[Pong]{
\includegraphics[scale=0.24]{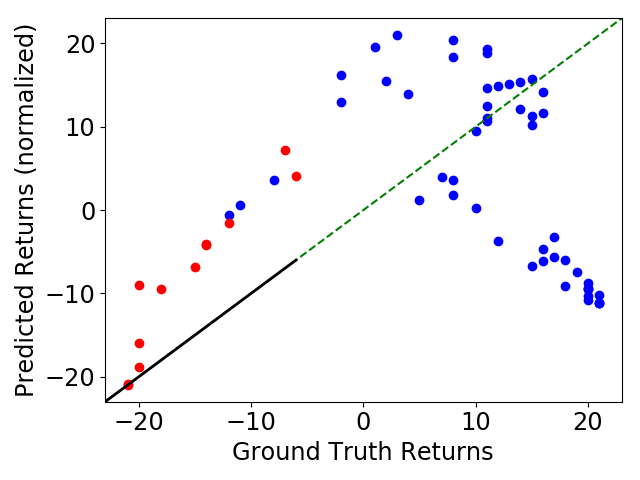}
\label{subfig:simpleCakmak3}
}
\subfigure[Q*bert]{
\includegraphics[scale=0.24]{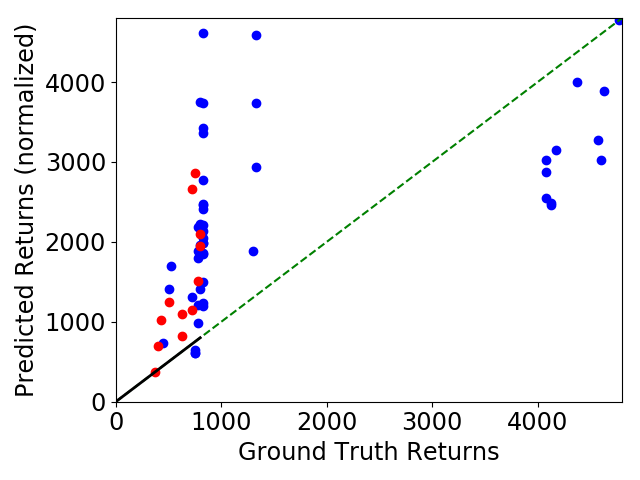}
\label{subfig:simpleCakmak4}
}
\subfigure[Seaquest]{
\includegraphics[scale=0.24]{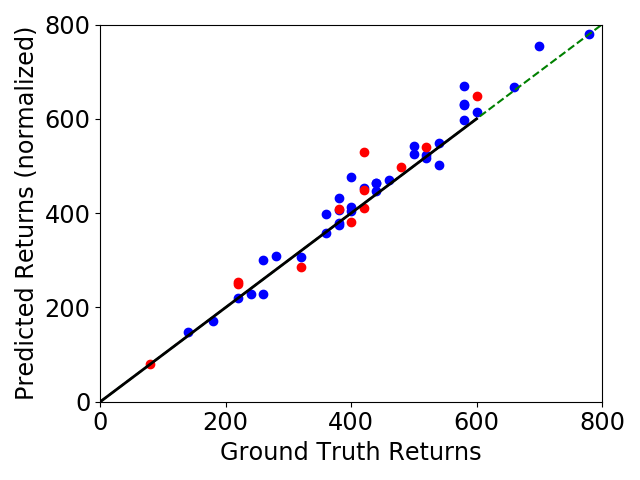}
\label{subfig:simpleCakmak5}
}
\subfigure[Space Invaders]{
\includegraphics[scale=0.24]{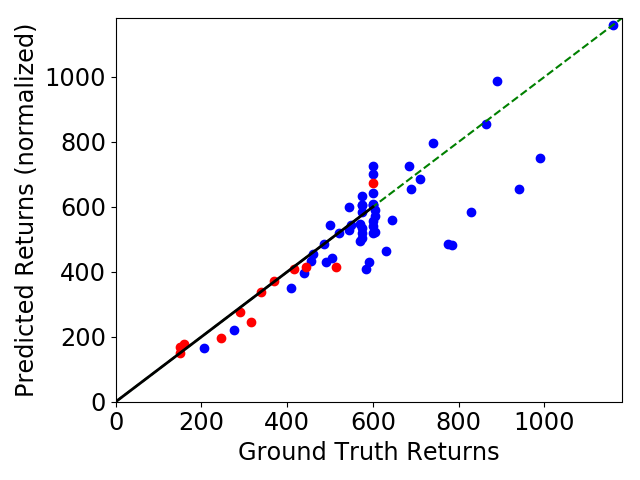}
\label{subfig:simpleCakmak6}
}
\caption{Extrapolation plots for Atari games. We compare ground truth returns over demonstrations to the predicted returns using T-REX (normalized to be in the same range as the ground truth returns). The black solid line represents the performance range of the demonstrator. The green dashed line represents extrapolation beyond the range of the demonstrator's performance.}
\label{fig:atari_extrapolation.}
\end{figure*}

\paragraph{Reward Extrapolation} We also examined the extrapolation of the reward function learned using T-REX. Results are shown in Figure~\ref{fig:atari_extrapolation.}. We observed accurate extrapolation for Beam Rider, Breakout, Enduro, Seaquest, and Space Invaders---five games where T-REX significantly outperform the demonstrator. The learned rewards for Pong, Q*bert, and Hero show less correlation. On Pong, T-REX overfits to the suboptimal demonstrations and ends up preferring longer games which do not result in a significant win or loss. T-REX is unable to score any points on Hero, likely due to poor extrapolation and the higher complexity of the game. Surprisingly, the learned reward function for Q*bert shows poor extrapolation, yet T-REX is able to outperform the demonstrator. We analyzed the resulting policy for Q*bert and found that PPO learns a repeatable way to score points by inducing Coily to jump off the edge. This behavior was not seen in the demonstrations. In the appendix, we plot the maximum and minimum predicted observations from the trajectories used to create Figure~\ref{fig:atari_extrapolation.} along with attention maps for the learned reward functions. 

 \subsubsection{Human Demonstrations}
 The above results used synthetic demonstrations generated from an RL agent. We also tested T-REX when given ground-truth rankings over human demonstrations. We used novice human demonstrations from the Atari Grand Challenge Dataset \cite{kurin2017atari} for five Atari tasks. T-REX was able to significantly outperform the best human demonstration in Q*bert, Space Invaders, and Video Pinball, but was unable to outperform the human in Montezuma's Revenge and Ms Pacman (see the appendix for details).


\subsection{Robustness to Noisy Rankings}

\begin{figure}[t]
    \centering
    \includegraphics[width=1.0\linewidth]{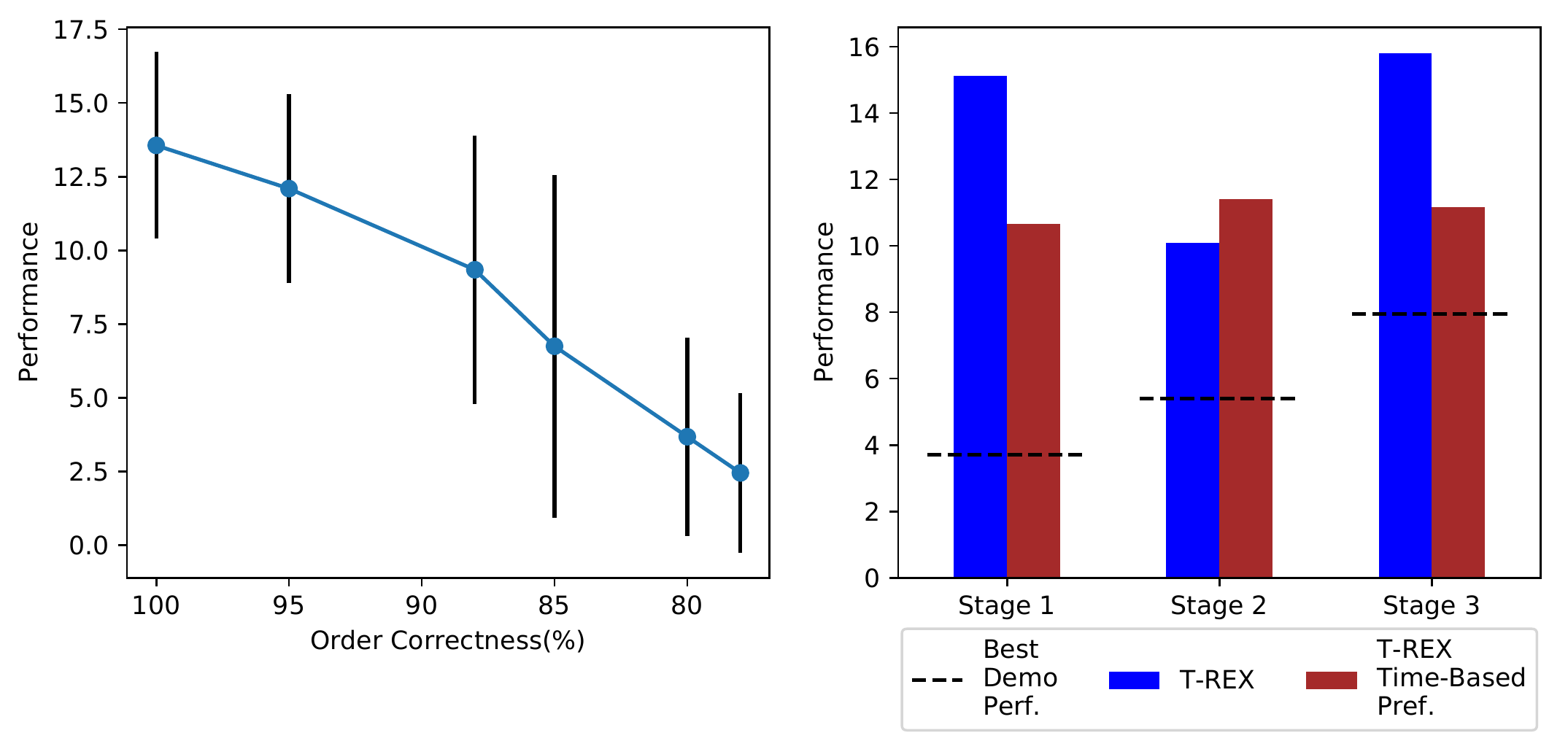}
    \vspace{-0.2in}
    \caption{(left): The performance of T-REX for different amounts of pairwise ranking noise in the Hopper domain. T-REX shows graceful degradation as ranking noise increases. The reward function is trained on stage-1 Hopper demonstrations. The graph shows the mean across 9 trials and 95\% confidence interval. (right): T-REX results with time-based rankings in the Hopper domain.}
    \label{fig:noise_level_anal}
\end{figure}

All experiments described thus far have had access to ground-truth rankings. To explore the effects of noisy rankings we first examined the stage 1 Hopper task.
We synthetically generated ranking noise by starting with a list of trajectories sorted by ground-truth returns and randomly swapping adjacent trajectories. By varying the number of swaps, we were able to generate different noise levels. Given $n$ trajectories in a ranked list provides ${n \choose 2}$ pairwise preferences over trajectories. The noise level is measured as a total order correctness: the fraction of trajectory pairs whose pairwise ranking after random swapping matches the original ground-truth pairwise preferences. The results of this experiment, averaged over 9 runs per noise level, are shown in Figure~\ref{fig:noise_level_anal}. We found that T-REX is relatively robust to noise of up to around 15\% pairwise errors. 

To examine the effect of noisy human rankings, we used the synthetic PPO demonstrations that were used in the previous Atari experiments and used Amazon Mechanical Turk to collect human rankings. We presented videos of the demonstrations in pairs along with a brief text description of the goal of the game and asked workers to select which demonstration had better performance, with an option for selecting ``Not Sure". We collected six labels per demonstration pair and used the most-common label as the label for training the reward function. We removed from the training data any pairings where there was a tie for the most-common label or where ``Not Sure" was the most common label. We found that despite this preprocessing step, human labels added a significant amount of noise and resulted in pair-wise rankings with accuracy between 63\% and 88\% when compared to ground-truth labels. However, despite significant ranking noise, T-REX outperformed the demonstrator on 5 of the 8 Atari games (see the appendix for full details).

\subsubsection{Learning from Time-Based Rankings}
Finally, we tested whether T-REX has the potential to work without explicit rankings. We took the same demonstrations used for the Mujoco tasks, and rather than sorting them based on ground-truth rankings, we used the order in which they were generated by PPO to produce a ranked list of trajectories, ordered by timestamp from earliest to latest. This provides ranked demonstrations without any need for demonstrator labels, and enables us to test whether simply observing an agent learn over time allows us to extrapolate intention by assuming that later trajectories are preferable to trajectories produced earlier in learning.
The results for Hopper are shown in Figure~\ref{fig:noise_level_anal} and other task results are shown in the appendix. We found that T-REX is able to infer a meaningful reward function even when noisy, time-based rankings are provided. All the trained policies produced comparable results on most stages to the ground-truth rankings, and those policies outperform BCO and GAIL on all tasks and stages except for Ant Stage 2.

\section{Conclusion}
In this paper, we introduced T-REX, a reward learning technique for high-dimensional tasks that can learn to extrapolate intent from suboptimal ranked demonstrations. To the best of our knowledge, this is the first IRL algorithm that is able to significantly outperform the demonstrator without additional external knowledge (e.g. signs of feature contributions to reward) and that scales to high-dimensional Atari games. 
When combined with deep reinforcement learning, we showed that this approach achieves better-than-demonstrator performance as well as outperforming state-of-the-art behavioral cloning and IRL methods. We also demonstrated that T-REX is robust to modest amounts of ranking noise, and can learn from automatically generated labels, obtained by watching a learner noisily improve at a task over time. 


\section*{Acknowledgments}
This  work  has  taken  place  in  the  Personal  AutonomousRobotics Lab (PeARL) at The University of Texas at Austin. PeARL  research  is  supported  in  part  by  the  NSF  (IIS-1724157, IIS-1638107, IIS-1617639, IIS-1749204) and ONR(N00014-18-2243).

{\small
\bibliography{lfd_refs}

\begin{thebibliography}{56}
\providecommand{\natexlab}[1]{#1}
\providecommand{\url}[1]{\texttt{#1}}
\expandafter\ifx\csname urlstyle\endcsname\relax
  \providecommand{\doi}[1]{doi: #1}\else
  \providecommand{\doi}{doi: \begingroup \urlstyle{rm}\Url}\fi

\bibitem[Abbeel \& Ng(2004)Abbeel and Ng]{abbeel2004apprenticeship}
Abbeel, P. and Ng, A.~Y.
\newblock Apprenticeship learning via inverse reinforcement learning.
\newblock In \emph{Proceedings of the 21st international conference on Machine
  learning}, 2004.

\bibitem[Akrour et~al.(2011)Akrour, Schoenauer, and
  Sebag]{akrour2011preference}
Akrour, R., Schoenauer, M., and Sebag, M.
\newblock Preference-based policy learning.
\newblock In \emph{Joint European Conference on Machine Learning and Knowledge
  Discovery in Databases}, pp.\  12--27. Springer, 2011.

\bibitem[Amodei et~al.(2016)Amodei, Olah, Steinhardt, Christiano, Schulman, and
  Man{\'e}]{amodei2016concrete}
Amodei, D., Olah, C., Steinhardt, J., Christiano, P., Schulman, J., and
  Man{\'e}, D.
\newblock Concrete problems in ai safety.
\newblock \emph{arXiv preprint arXiv:1606.06565}, 2016.

\bibitem[Argall et~al.(2009)Argall, Chernova, Veloso, and Browning]{Argall2009}
Argall, B.~D., Chernova, S., Veloso, M., and Browning, B.
\newblock A survey of robot learning from demonstration.
\newblock \emph{Robotics and autonomous systems}, 57\penalty0 (5):\penalty0
  469--483, 2009.

\bibitem[Arora \& Doshi(2018)Arora and Doshi]{arora2018survey}
Arora, S. and Doshi, P.
\newblock A survey of inverse reinforcement learning: Challenges, methods and
  progress.
\newblock \emph{arXiv preprint arXiv:1806.06877}, 2018.

\bibitem[Aytar et~al.(2018)Aytar, Pfaff, Budden, Paine, Wang, and
  de~Freitas]{imitationyoutube}
Aytar, Y., Pfaff, T., Budden, D., Paine, T.~L., Wang, Z., and de~Freitas, N.
\newblock Playing hard exploration games by watching youtube.
\newblock \emph{arXiv preprint arXiv:1805.11592}, 2018.

\bibitem[Boularias et~al.(2011)Boularias, Kober, and
  Peters]{boularias2011relative}
Boularias, A., Kober, J., and Peters, J.
\newblock Relative entropy inverse reinforcement learning.
\newblock In \emph{Proceedings of the Fourteenth International Conference on
  Artificial Intelligence and Statistics}, pp.\  182--189, 2011.

\bibitem[Bradley \& Terry(1952)Bradley and Terry]{bradley1952rank}
Bradley, R.~A. and Terry, M.~E.
\newblock Rank analysis of incomplete block designs: I. the method of paired
  comparisons.
\newblock \emph{Biometrika}, 39\penalty0 (3/4):\penalty0 324--345, 1952.

\bibitem[Brockman et~al.(2016)Brockman, Cheung, Pettersson, Schneider,
  Schulman, Tang, and Zaremba]{brockman2016openai}
Brockman, G., Cheung, V., Pettersson, L., Schneider, J., Schulman, J., Tang,
  J., and Zaremba, W.
\newblock Openai gym.
\newblock \emph{arXiv preprint arXiv:1606.01540}, 2016.

\bibitem[Burchfiel et~al.(2016)Burchfiel, Tomasi, and Parr]{dmirl}
Burchfiel, B., Tomasi, C., and Parr, R.
\newblock Distance minimization for reward learning from scored trajectories.
\newblock In \emph{AAAI}, pp.\  3330--3336, 2016.

\bibitem[Choi et~al.(2019)Choi, Lee, and Oh]{choi2019robust}
Choi, S., Lee, K., and Oh, S.
\newblock Robust learning from demonstrations with mixed qualities using
  leveraged gaussian processes.
\newblock \emph{IEEE Transactions on Robotics}, 2019.

\bibitem[Christiano et~al.(2017)Christiano, Leike, Brown, Martic, Legg, and
  Amodei]{christiano2017deep}
Christiano, P.~F., Leike, J., Brown, T., Martic, M., Legg, S., and Amodei, D.
\newblock Deep reinforcement learning from human preferences.
\newblock In \emph{Advances in Neural Information Processing Systems}, pp.\
  4299--4307, 2017.

\bibitem[Dhariwal et~al.(2017)Dhariwal, Hesse, Klimov, Nichol, Plappert,
  Radford, Schulman, Sidor, Wu, and Zhokhov]{baselines}
Dhariwal, P., Hesse, C., Klimov, O., Nichol, A., Plappert, M., Radford, A.,
  Schulman, J., Sidor, S., Wu, Y., and Zhokhov, P.
\newblock Openai baselines.
\newblock \url{https://github.com/openai/baselines}, 2017.

\bibitem[El~Asri et~al.(2016)El~Asri, Piot, Geist, Laroche, and
  Pietquin]{scorebasedirl}
El~Asri, L., Piot, B., Geist, M., Laroche, R., and Pietquin, O.
\newblock Score-based inverse reinforcement learning.
\newblock In \emph{Proceedings of the 2016 International Conference on
  Autonomous Agents \& Multiagent Systems}, pp.\  457--465. International
  Foundation for Autonomous Agents and Multiagent Systems, 2016.

\bibitem[Finn et~al.(2016)Finn, Levine, and Abbeel]{finn2016guided}
Finn, C., Levine, S., and Abbeel, P.
\newblock Guided cost learning: Deep inverse optimal control via policy
  optimization.
\newblock In \emph{International Conference on Machine Learning}, 2016.

\bibitem[Fu et~al.(2017)Fu, Luo, and Levine]{airl}
Fu, J., Luo, K., and Levine, S.
\newblock Learning robust rewards with adversarial inverse reinforcement
  learning.
\newblock \emph{arXiv preprint arXiv:1710.11248}, 2017.

\bibitem[Gao et~al.(2012)Gao, Peters, Tsourdos, Zhifei, and
  Meng~Joo]{gao2012survey}
Gao, Y., Peters, J., Tsourdos, A., Zhifei, S., and Meng~Joo, E.
\newblock A survey of inverse reinforcement learning techniques.
\newblock \emph{International Journal of Intelligent Computing and
  Cybernetics}, 5\penalty0 (3):\penalty0 293--311, 2012.

\bibitem[Gao et~al.(2018)Gao, Lin, Yu, Levine, Darrell,
  et~al.]{gao2018reinforcement}
Gao, Y., Lin, J., Yu, F., Levine, S., Darrell, T., et~al.
\newblock Reinforcement learning from imperfect demonstrations.
\newblock \emph{arXiv preprint arXiv:1802.05313}, 2018.

\bibitem[Goo \& Niekum(2019)Goo and Niekum]{goo2019one}
Goo, W. and Niekum, S.
\newblock One-shot learning of multi-step tasks from observation via activity
  localization in auxiliary video.
\newblock In \emph{2019 IEEE International Conference on Robotics and
  Automation (ICRA)}, 2019.

\bibitem[Goodfellow et~al.(2014)Goodfellow, Pouget-Abadie, Mirza, Xu,
  Warde-Farley, Ozair, Courville, and Bengio]{goodfellow2014generative}
Goodfellow, I., Pouget-Abadie, J., Mirza, M., Xu, B., Warde-Farley, D., Ozair,
  S., Courville, A., and Bengio, Y.
\newblock Generative adversarial nets.
\newblock In \emph{Advances in neural information processing systems}, pp.\
  2672--2680, 2014.

\bibitem[Greydanus et~al.(2018)Greydanus, Koul, Dodge, and
  Fern]{greydanus2018visualizing}
Greydanus, S., Koul, A., Dodge, J., and Fern, A.
\newblock Visualizing and understanding atari agents.
\newblock In \emph{International Conference on Machine Learning}, pp.\
  1787--1796, 2018.

\bibitem[Grollman \& Billard(2011)Grollman and Billard]{grollman2011donut}
Grollman, D.~H. and Billard, A.
\newblock Donut as i do: Learning from failed demonstrations.
\newblock In \emph{Robotics and Automation (ICRA), 2011 IEEE International
  Conference on}, pp.\  3804--3809. IEEE, 2011.

\bibitem[Henderson et~al.(2018)Henderson, Chang, Bacon, Meger, Pineau, and
  Precup]{henderson2018optiongan}
Henderson, P., Chang, W.-D., Bacon, P.-L., Meger, D., Pineau, J., and Precup,
  D.
\newblock Optiongan: Learning joint reward-policy options using generative
  adversarial inverse reinforcement learning.
\newblock In \emph{Thirty-Second AAAI Conference on Artificial Intelligence},
  2018.

\bibitem[Hester et~al.(2017)Hester, Vecerik, Pietquin, Lanctot, Schaul, Piot,
  Horgan, Quan, Sendonaris, Dulac-Arnold, et~al.]{hester2017deep}
Hester, T., Vecerik, M., Pietquin, O., Lanctot, M., Schaul, T., Piot, B.,
  Horgan, D., Quan, J., Sendonaris, A., Dulac-Arnold, G., et~al.
\newblock Deep q-learning from demonstrations.
\newblock \emph{arXiv preprint arXiv:1704.03732}, 2017.

\bibitem[Ho \& Ermon(2016)Ho and Ermon]{ho2016generative}
Ho, J. and Ermon, S.
\newblock Generative adversarial imitation learning.
\newblock In \emph{Advances in Neural Information Processing Systems}, pp.\
  4565--4573, 2016.

\bibitem[Ibarz et~al.(2018)Ibarz, Leike, Pohlen, Irving, Legg, and
  Amodei]{ibarz2018reward}
Ibarz, B., Leike, J., Pohlen, T., Irving, G., Legg, S., and Amodei, D.
\newblock Reward learning from human preferences and demonstrations in atari.
\newblock In \emph{Advances in Neural Information Processing Systems}, pp.\
  8022--8034, 2018.

\bibitem[Kingma \& Ba(2014)Kingma and Ba]{kingma2014adam}
Kingma, D.~P. and Ba, J.
\newblock Adam: A method for stochastic optimization.
\newblock \emph{arXiv preprint arXiv:1412.6980}, 2014.

\bibitem[Kober \& Peters(2009)Kober and Peters]{kober2009policy}
Kober, J. and Peters, J.~R.
\newblock Policy search for motor primitives in robotics.
\newblock In \emph{Advances in neural information processing systems}, pp.\
  849--856, 2009.

\bibitem[Kurin et~al.(2017)Kurin, Nowozin, Hofmann, Beyer, and
  Leibe]{kurin2017atari}
Kurin, V., Nowozin, S., Hofmann, K., Beyer, L., and Leibe, B.
\newblock The atari grand challenge dataset.
\newblock \emph{arXiv preprint arXiv:1705.10998}, 2017.

\bibitem[Liu et~al.(2018)Liu, Gupta, Abbeel, and Levine]{liu2018imitation}
Liu, Y., Gupta, A., Abbeel, P., and Levine, S.
\newblock Imitation from observation: Learning to imitate behaviors from raw
  video via context translation.
\newblock In \emph{2018 IEEE International Conference on Robotics and
  Automation (ICRA)}, pp.\  1118--1125. IEEE, 2018.

\bibitem[Luce(2012)]{luce2012individual}
Luce, R.~D.
\newblock \emph{Individual choice behavior: A theoretical analysis}.
\newblock Courier Corporation, 2012.

\bibitem[Mnih et~al.(2015)Mnih, Kavukcuoglu, Silver, Rusu, Veness, Bellemare,
  Graves, Riedmiller, Fidjeland, Ostrovski, et~al.]{dqn}
Mnih, V., Kavukcuoglu, K., Silver, D., Rusu, A.~A., Veness, J., Bellemare,
  M.~G., Graves, A., Riedmiller, M., Fidjeland, A.~K., Ostrovski, G., et~al.
\newblock Human-level control through deep reinforcement learning.
\newblock \emph{Nature}, 518\penalty0 (7540):\penalty0 529, 2015.

\bibitem[Ng et~al.(1999)Ng, Harada, and Russell]{ng1999policy}
Ng, A.~Y., Harada, D., and Russell, S.
\newblock Policy invariance under reward transformations: Theory and
  application to reward shaping.
\newblock In \emph{ICML}, volume~99, pp.\  278--287, 1999.

\bibitem[Osa et~al.(2018)Osa, Pajarinen, Neumann, Bagnell, Abbeel, Peters,
  et~al.]{osa2018algorithmic}
Osa, T., Pajarinen, J., Neumann, G., Bagnell, J.~A., Abbeel, P., Peters, J.,
  et~al.
\newblock An algorithmic perspective on imitation learning.
\newblock \emph{Foundations and Trends{\textregistered} in Robotics},
  7\penalty0 (1-2):\penalty0 1--179, 2018.

\bibitem[Pomerleau(1991)]{pomerleau1991efficient}
Pomerleau, D.~A.
\newblock Efficient training of artificial neural networks for autonomous
  navigation.
\newblock \emph{Neural Computation}, 3\penalty0 (1):\penalty0 88--97, 1991.

\bibitem[Puterman(2014)]{puterman2014markov}
Puterman, M.~L.
\newblock \emph{Markov decision processes: discrete stochastic dynamic
  programming}.
\newblock John Wiley \& Sons, 2014.

\bibitem[Qureshi \& Yip(2018)Qureshi and Yip]{qureshi2018adversarial}
Qureshi, A.~H. and Yip, M.~C.
\newblock Adversarial imitation via variational inverse reinforcement learning.
\newblock \emph{arXiv preprint arXiv:1809.06404}, 2018.

\bibitem[Ramachandran \& Amir(2007)Ramachandran and
  Amir]{ramachandran2007bayesian}
Ramachandran, D. and Amir, E.
\newblock Bayesian inverse reinforcement learning.
\newblock In \emph{Proceedings of the 20th International Joint Conference on
  Artifical intelligence}, pp.\  2586--2591, 2007.

\bibitem[Ross et~al.(2011)Ross, Gordon, and Bagnell]{ross2011reduction}
Ross, S., Gordon, G., and Bagnell, D.
\newblock A reduction of imitation learning and structured prediction to
  no-regret online learning.
\newblock In \emph{Proceedings of the fourteenth international conference on
  artificial intelligence and statistics}, pp.\  627--635, 2011.

\bibitem[Schulman et~al.(2017)Schulman, Wolski, Dhariwal, Radford, and
  Klimov]{schulman2017proximal}
Schulman, J., Wolski, F., Dhariwal, P., Radford, A., and Klimov, O.
\newblock Proximal policy optimization algorithms.
\newblock \emph{arXiv preprint arXiv:1707.06347}, 2017.

\bibitem[Sermanet et~al.(2018)Sermanet, Lynch, Chebotar, Hsu, Jang, Schaal,
  Levine, and Brain]{sermanet2018time}
Sermanet, P., Lynch, C., Chebotar, Y., Hsu, J., Jang, E., Schaal, S., Levine,
  S., and Brain, G.
\newblock Time-contrastive networks: Self-supervised learning from video.
\newblock In \emph{2018 IEEE International Conference on Robotics and
  Automation (ICRA)}, pp.\  1134--1141. IEEE, 2018.

\bibitem[Shiarlis et~al.(2016)Shiarlis, Messias, and
  Whiteson]{shiarlis2016inverse}
Shiarlis, K., Messias, J., and Whiteson, S.
\newblock Inverse reinforcement learning from failure.
\newblock In \emph{Proceedings of the 2016 International Conference on
  Autonomous Agents \& Multiagent Systems}, pp.\  1060--1068. International
  Foundation for Autonomous Agents and Multiagent Systems, 2016.

\bibitem[Sugiyama et~al.(2012)Sugiyama, Meguro, and Minami]{irldialog}
Sugiyama, H., Meguro, T., and Minami, Y.
\newblock Preference-learning based inverse reinforcement learning for dialog
  control.
\newblock In \emph{INTERSPEECH}, pp.\  222--225, 2012.

\bibitem[Sutton \& Barto(1998)Sutton and Barto]{sutton1998introduction}
Sutton, R.~S. and Barto, A.~G.
\newblock \emph{Introduction to reinforcement learning}, volume 135.
\newblock MIT press Cambridge, 1998.

\bibitem[Syed \& Schapire(2008)Syed and Schapire]{syed2008game}
Syed, U. and Schapire, R.~E.
\newblock A game-theoretic approach to apprenticeship learning.
\newblock In \emph{Advances in neural information processing systems}, pp.\
  1449--1456, 2008.

\bibitem[Taylor et~al.(2011)Taylor, Suay, and Chernova]{taylor2011integrating}
Taylor, M.~E., Suay, H.~B., and Chernova, S.
\newblock Integrating reinforcement learning with human demonstrations of
  varying ability.
\newblock In \emph{The 10th International Conference on Autonomous Agents and
  Multiagent Systems-Volume 2}, pp.\  617--624. International Foundation for
  Autonomous Agents and Multiagent Systems, 2011.

\bibitem[Todorov et~al.(2012)Todorov, Erez, and Tassa]{todorov2012mujoco}
Todorov, E., Erez, T., and Tassa, Y.
\newblock Mujoco: A physics engine for model-based control.
\newblock In \emph{Intelligent Robots and Systems (IROS), 2012 IEEE/RSJ
  International Conference on}, pp.\  5026--5033. IEEE, 2012.

\bibitem[Torabi et~al.(2018{\natexlab{a}})Torabi, Warnell, and
  Stone]{torabi2018behavioral}
Torabi, F., Warnell, G., and Stone, P.
\newblock Behavioral cloning from observation.
\newblock In \emph{Proceedings of the 27th International Joint Conference on
  Artificial Intelligence (IJCAI)}, July 2018{\natexlab{a}}.

\bibitem[Torabi et~al.(2018{\natexlab{b}})Torabi, Warnell, and
  Stone]{torabi2018generative}
Torabi, F., Warnell, G., and Stone, P.
\newblock Generative adversarial imitation from observation.
\newblock \emph{arXiv preprint arXiv:1807.06158}, 2018{\natexlab{b}}.

\bibitem[Tucker et~al.(2018)Tucker, Gleave, and Russell]{irlvideogames}
Tucker, A., Gleave, A., and Russell, S.
\newblock Inverse reinforcement learning for video games.
\newblock In \emph{Proceedings of the Workshop on Deep Reinforcement Learning
  at NeurIPS}, 2018.

\bibitem[Wirth et~al.(2016)Wirth, F{\"u}rnkranz, and Neumann]{pbirl}
Wirth, C., F{\"u}rnkranz, J., and Neumann, G.
\newblock Model-free preference-based reinforcement learning.
\newblock In \emph{Proceedings of the Thirtieth AAAI Conference on Artificial
  Intelligence}, pp.\  2222--2228. AAAI Press, 2016.

\bibitem[Wirth et~al.(2017)Wirth, Akrour, Neumann, and
  F{{{\"u}}}rnkranz]{preferencesurvey}
Wirth, C., Akrour, R., Neumann, G., and F{{{\"u}}}rnkranz, J.
\newblock A survey of preference-based reinforcement learning methods.
\newblock \emph{Journal of Machine Learning Research}, 18\penalty0
  (136):\penalty0 1--46, 2017.
\newblock URL \url{http://jmlr.org/papers/v18/16-634.html}.

\bibitem[Wulfmeier et~al.(2015)Wulfmeier, Ondruska, and
  Posner]{wulfmeier2015maximum}
Wulfmeier, M., Ondruska, P., and Posner, I.
\newblock Maximum entropy deep inverse reinforcement learning.
\newblock \emph{arXiv preprint arXiv:1507.04888}, 2015.

\bibitem[Yu et~al.(2018)Yu, Finn, Xie, Dasari, Zhang, Abbeel, and
  Levine]{yu2018one}
Yu, T., Finn, C., Xie, A., Dasari, S., Zhang, T., Abbeel, P., and Levine, S.
\newblock One-shot imitation from observing humans via domain-adaptive
  meta-learning.
\newblock \emph{arXiv preprint arXiv:1802.01557}, 2018.

\bibitem[Zheng et~al.(2014)Zheng, Liu, and Ni]{zheng2014robust}
Zheng, J., Liu, S., and Ni, L.~M.
\newblock Robust bayesian inverse reinforcement learning with sparse behavior
  noise.
\newblock In \emph{Proceedings of the AAAI Conference on Artificial
  Intelligence}, pp.\  2198--2205, 2014.

\bibitem[Ziebart et~al.(2008)Ziebart, Maas, Bagnell, and
  Dey]{ziebart2008maximum}
Ziebart, B.~D., Maas, A.~L., Bagnell, J.~A., and Dey, A.~K.
\newblock Maximum entropy inverse reinforcement learning.
\newblock In \emph{Proceedings of the 23rd AAAI Conference on Artificial
  Intelligence}, pp.\  1433--1438, 2008.

\end{thebibliography}
\bibliographystyle{icml2019}
}

\newpage

\appendix
\setcounter{figure}{0}
\setcounter{table}{0}

\section{Code and Videos}
Code as well as supplemental videos are available at \url{https://github.com/hiwonjoon/ICML2019-TREX}.

\section{T-REX Results on the MuJoCo Domain}

\subsection{Policy performance}
Table~\ref{table:mujoco_rl_results} shows the full results for the MuJoCo experiments. The T-REX (time-ordered) row shows the resulting performance of T-REX when demonstrations come from observing a learning agent and are ranked based on timestamps rather than using explicit preference rankings.

\begin{table*}[t]
\caption{The results on three robotic locomotion tasks when given suboptimal demonstrations. Performance is measured as the total distance traveled, as measured by the final x-position of the robot's body. For each stage and task, the best performance given suboptimal demonstrations is shown on the top row, and the best achievable performance (i.e. performance achieved by a PPO agent) under the ground-truth reward is shown on the bottom row. The mean and standard deviation are based on 25 trials (obtained by running PPO five times and for each run of PPO performing five policy rollouts). The first row of T-REX results show the performance when demonstrations are ranked using the ground-truth returns. The second row of T-REX shows results for learning from observing a learning agent (time-ordered). The demonstrations are ranked based on the timestamp when they were produced by the PPO algorithm learning to perform the task.}
\label{table:mujoco_rl_results}
\vskip 0.15in
\centering
\begin{tabular}{c|ccc|ccc|cc}
\toprule
& \multicolumn{3}{c|}{HalfCheetah}                                                                                                                                          & \multicolumn{3}{c|}{Hopper}                                                                                                                                               & \multicolumn{2}{c}{Ant}                                                                                         \\ \hline
                                                            & Stage 1                                                & Stage 2                                                & Stage 3                                                 & Stage 1                                                & \multicolumn{1}{l}{Stage 2}                             & Stage 3                                                & Stage 1                                                & Stage 2                                                 \\ \hline
\begin{tabular}[c]{@{}c@{}}Best Demo\\ Performance\end{tabular}  & \begin{tabular}[c]{@{}c@{}}12.52\\ (1.04)\end{tabular} & \begin{tabular}[c]{@{}c@{}}44.98\\ (0.60)\end{tabular} & \begin{tabular}[c]{@{}c@{}}89.87\\ (8.15)\end{tabular}  & \begin{tabular}[c]{@{}c@{}}3.70\\ (0.01)\end{tabular}  & \begin{tabular}[c]{@{}c@{}}5.40\\ (0.12)\end{tabular}   & \begin{tabular}[c]{@{}c@{}}7.95\\ (1.64)\end{tabular}  & \begin{tabular}[c]{@{}c@{}}1.56\\ (1.28)\end{tabular}  & \begin{tabular}[c]{@{}c@{}}\textbf{54.64}\\ (22.09)\end{tabular} \\ \hline

\begin{tabular}[c]{@{}c@{}}T-REX\\ (ours)\end{tabular}      & \begin{tabular}[c]{@{}c@{}}46.90\\ (1.89)\end{tabular} & \begin{tabular}[c]{@{}c@{}}\textbf{61.56}\\ (10.96)\end{tabular} & \begin{tabular}[c]{@{}c@{}}143.40\\ (3.84)\end{tabular} & \begin{tabular}[c]{@{}c@{}}\textbf{15.13}\\ (3.21)\end{tabular} & \begin{tabular}[c]{@{}c@{}}10.10\\ (1.68)\end{tabular}  & \begin{tabular}[c]{@{}c@{}}\textbf{15.80}\\ (0.37)\end{tabular} & \begin{tabular}[c]{@{}c@{}}4.93\\ (2.86)\end{tabular} & \begin{tabular}[c]{@{}c@{}}7.34\\ (2.50)\end{tabular}  \\

\begin{tabular}[c]{@{}c@{}}T-REX\\ (time-ordered)\end{tabular}  & \begin{tabular}[c]{@{}c@{}}\textbf{51.39}\\(4.52)\end{tabular}  & \begin{tabular}[c]{@{}c@{}}54.90\\(2.29)\end{tabular} & \begin{tabular}[c]{@{}c@{}}\textbf{154.67}\\(57.43)\end{tabular} & \begin{tabular}[c]{@{}c@{}}10.66\\(3.76)\end{tabular}  & \begin{tabular}[c]{@{}c@{}}\textbf{11.41}\\(0.56)\end{tabular} & \begin{tabular}[c]{@{}c@{}}11.17\\(0.60)\end{tabular}  & \begin{tabular}[c]{@{}c@{}}\textbf{5.55}\\(5.86)\end{tabular}  & \begin{tabular}[c]{@{}c@{}}1.28\\(0.28)\end{tabular} \\

\begin{tabular}[c]{@{}c@{}}BCO\end{tabular}  & \begin{tabular}[c]{@{}c@{}}7.71\\(8.35)\end{tabular}  & \begin{tabular}[c]{@{}c@{}}23.59\\(8.33)\end{tabular} & \begin{tabular}[c]{@{}c@{}}57.13\\(19.14)\end{tabular} & \begin{tabular}[c]{@{}c@{}}3.52\\(0.14)\end{tabular}  & \begin{tabular}[c]{@{}c@{}}4.41\\(1.45)\end{tabular} & \begin{tabular}[c]{@{}c@{}}4.58\\(1.07)\end{tabular}  & \begin{tabular}[c]{@{}c@{}}1.06\\(1.79)\end{tabular}  & \begin{tabular}[c]{@{}c@{}}26.56\\(12.96)\end{tabular} \\

\begin{tabular}[c]{@{}c@{}}GAIL\end{tabular}  & \begin{tabular}[c]{@{}c@{}}7.39\\(4.12)\end{tabular}  & \begin{tabular}[c]{@{}c@{}}8.42\\(3.43)\end{tabular} & \begin{tabular}[c]{@{}c@{}}26.28\\(12.73)\end{tabular} & \begin{tabular}[c]{@{}c@{}}8.09\\(3.25)\end{tabular}  & \begin{tabular}[c]{@{}c@{}}10.99\\(2.35)\end{tabular} & \begin{tabular}[c]{@{}c@{}}12.63\\(3.66)\end{tabular}  & \begin{tabular}[c]{@{}c@{}}0.95\\(2.06)\end{tabular}  & \begin{tabular}[c]{@{}c@{}}5.84\\(4.08)\end{tabular} \\


\hline
\begin{tabular}[c]{@{}c@{}}Best w/\\ GT Reward\end{tabular} & \multicolumn{3}{c|}{\begin{tabular}[c]{@{}c@{}}199.11\\ (9.08)\end{tabular}}                                                                                              & \multicolumn{3}{c|}{\begin{tabular}[c]{@{}c@{}}15.94\\ (1.47)\end{tabular}}                                                                                               & \multicolumn{2}{c}{\begin{tabular}[c]{@{}c@{}}182.23\\ (8.98)\end{tabular}}\\                 
\bottomrule
\end{tabular}
\end{table*}

\subsection{Policy visualization}

\begin{figure*}
    \centering
    \includegraphics[width=0.99\linewidth]{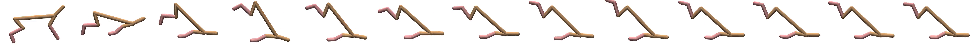}
    \includegraphics[width=0.99\linewidth]{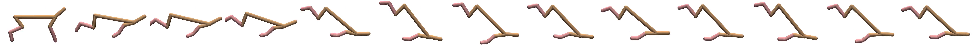}
    (a) Stage 1
    \includegraphics[width=0.99\linewidth]{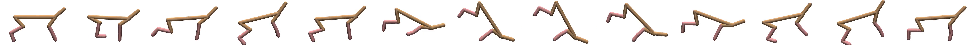}
    \includegraphics[width=0.99\linewidth]{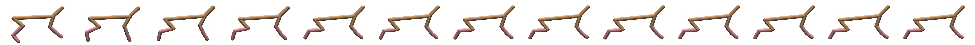}
    (b) Stage 2
    \includegraphics[width=0.99\linewidth]{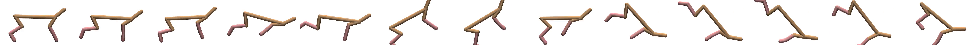}
    \includegraphics[width=0.99\linewidth]{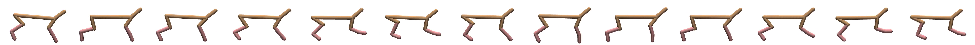}
    (c) Stage 3\\
    \caption{HalfCheetah policy visualization. For each subplot, (top) is the best given demonstration policy in a stage, and (bottom) is the trained policy with a T-REX reward function.}
    \label{fig:cheetah_policy}
\end{figure*}

We visualized the T-REX-learned policy for HalfCheetah in Figure~\ref{fig:cheetah_policy}. Visualizing the demonstrations from different stages shows the specific way the policy evolves over time; an agent learns to crawl first and then begins to attempt to walk in an upright position. The T-REX policy learned from the highly suboptimal Stage 1 demonstrations results in a similar-style crawling gait; however, T-REX captures some of the intent behind the demonstration and is able to optimize a gait that resembles the demonstrator but with increased speed, resulting in a better-than-demonstrator policy. Similarly, given demonstrations from Stage 2, which are still highly suboptimal, T-REX learns a policy that resembles the gait of the best demonstration, but is able to optimize and partially stabilize this gait. Finally, given demonstrations from Stage 3, which are still suboptimal, T-REX is able to learn a near-optimal gait. 

\section{Behavioral Cloning from Observation}\label{sec:bcoAppendix}
To build the inverse transition models used by BCO \cite{torabi2018behavioral} we used 20,000 steps of a random policy to collect transitions with labeled states. We used the Adam optimizer with learning rate 0.0001 and L2 regularization of 0.0001. We used the DQN architecture \cite{dqn} for the classification network, using the same architecture to predict actions given state transitions as well as predict actions given states. When predicting $P(a| s_t, s_{t+1})$, we concatenate the state vectors obtaining an 8x84x84 input consisting of two 4x84x84 frames representing $s_t$ and $s_{t+1}$.

We give both T-REX and BCO the full set of demonstrations. We tried to improve the performance of BCO by running behavioral cloning only on the best $X\%$ of the demonstrations, but were unable to find a parameter setting that performed better than $X=100$, likely due to a lack of training data when using very few demonstrations.

\section{Atari reward learning details}
We used the OpenAI Baselines implementation of PPO with default hyperparameters. We ran all of our experiments on an NVIDIA TITAN V GPU. We used 9 parallel workers when running PPO.

When learning and predicting rewards, we mask the score and number of lives left for all games. We did this to avoid having the network learn to only look at the score and recognize, say, the number of significant digits, etc. We additionally masked the sector number and number of enemy ships left on Beam Rider. We masked the bottom half of the dashboard for Enduro to mask the position of the car in the race. We masked the number of divers found and the oxygen meter for Seaquest. We masked the power level and inventory for Hero.

To train the reward network for Enduro, we randomly downsampled full trajectories. To create a training set we repeatedly randomly select two full demonstrations, then randomly cropped between 0 and 5 of the initial frames from each trajectory and then downsampled both trajectories by only keeping every $x$th frame where $x$ is randomly chosen between 3 and 6. We selected 2,000 randomly downsampled demonstrations and trained the reward network for 10,000 steps of Adam with a learning rate of 5e-5.

\clearpage

\section{Comparison to active reward learning}
In this section, we examine the ability of prior work on active preference learning to exceed the performance of the demonstrator. In Table~\ref{tab:democomparison}, we denote the results that surpass the best demonstration with an asterisk (*). DQfD+A only surpasses the demonstrator in 3 out of 9 games tested, even with thousands of active queries. Note that DQfD+A extends the original DQfD algorithm \cite{hester2017deep}, which uses demonstrations combined with RL on ground-truth rewards, yet is only able to surpass the best demonstration in 14 out of 41 Atari games. In contrast, we are able to leverage only 12 ranked demos to achieve better-than-demonstrator performance on 7 out of 8 games tested, without requiring access to true rewards or access to thousands of active queries from an oracle.

\citet{ibarz2018reward} combine Deep Q-learning from demonstrations and active preference queries (DQfD+A). DQfD+A uses demonstrations consisting of $(s_t,a_t,s_{t+1})$-tuples to initialize a policy using DQfD \cite{hester2017deep}. The algorithm then uses the active preference learning algorithm of \citet{christiano2017deep} to refine the inferred reward function and initial policy learned from demonstrations. The first two columns of Table~\ref{tab:democomparison} compare the demonstration quality given to DQfD+A and T-REX. While our results make use of more demonstrations (12 for T-REX versus 4--7 for DQfD+A), our demonstrations are typically orders of magnitude worse than the demonstrations used by DQfD+A: on average the demonstrations given to DQfD+A are 38 times better than those used by T-REX. However, despite this large gap in the performance of the demonstrations, T-REX surpasses the performance of DQfD+A on Q*Bert, and Seaquest. We achieve these results using 12 ranked demonstrations. This requires only 66 comparisons ($n\cdot (n-1) /2$) by the demonstrator. In comparison, the DQfD+A results used 3,400 preference labels obtained during policy training using ground-truth rewards.

\newpage
 \section{Human Demonstrations and Rankings}
\subsection{Human demonstrations}
We used the Atari Grand Challenge data set \cite{kurin2017atari} to collect actual human demonstrations for five Atari games. We used the ground truth returns in the Atari Grand Challenge data set to rank demonstrations. To generate demonstrations we removed duplicate demonstrations (human demonstrations that achieved the same score). We then sorted the remaining demonstrations based on ground truth return and selected 12 of these demonstrations to form our training set. We ran T-REX using the same hyperparameters as described above. 

The resulting performance of T-REX is shown in Table~\ref{tab:agc_trex}. T-REX is able to outperform the best human demonstration on Q*bert, Space Invaders, and Video Pinball; however, it is not able to learn a good control policy for Montezuma's Revenge or Ms Pacman. These games require maze navigation and balancing different objectives, such as collecting objects and avoiding enemies. This matches our results in the main text that show that T-REX is unable to learn a policy for playing Hero, a similar maze navigation task with multiple objectives such as blowing up walls, rescuing people, and destroying enemies. Extending T-REX to work in these types of settings is an interesting area of future work.

\subsection{Human rankings}
To measure the effects of human ranking noise, we took the same 12 PPO demonstrations described above in the main text and had humans rank the demonstrations. We used Amazon Mechanical Turk and showed the workers two side-by-side demonstrations and asked them to classify whether video A or video B had better performance or whether they were unsure. 

We took all 132 possible sequences of two videos across the 12 demonstrations and collected 6 labels for each pair of demonstrations. Because the workers are not actually giving the demonstrations and because some workers may exploit the task by simply selecting choices at random, we expect these labels to be a worst-case lower bound on the accuracy. To ameliorate the noise in the labels we take all 6 labels per pair and use the majority vote as the human label. If there is no majority or if the majority selects the ``Not Sure" label, then we do not include this pair in our training data for T-REX.


The resulting accuracy and number of labels that had a majority preference are shown in Table~\ref{tab:atariLearningFromHumanRankings}. We ran T-REX using the same hyperparameters described in the main text. We ran PPO with 3 different seeds and report the performance of the best final policy averaged over 30 trials. We found that surprisingly, T-REX is able to optimize good policies for many of the games, despite noisy labels. However, we did find cases such as Enduro, where the labels were too noisy to allow successful policy learning.

 \section{Atari Reward Visualizations}
\begin{figure*}
    \centering
    \subfigure[Beam Rider observation with maximum predicted reward]{
        \includegraphics[width=0.9\linewidth]{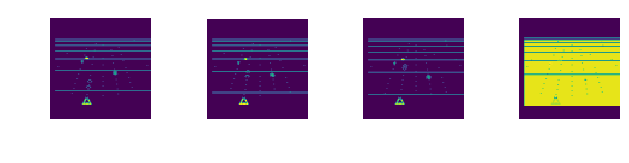}
        
    }
     \subfigure[Beam Rider reward model attention on maximum predicted reward]{
        \includegraphics[width=0.9\linewidth]{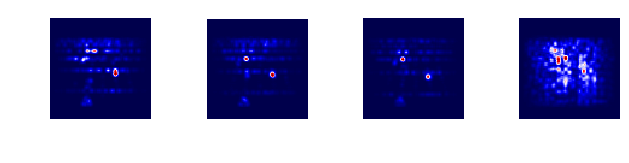}
        
    }
    \subfigure[Beam Rider observation with minimum predicted reward]{
        \includegraphics[width=0.9\linewidth]{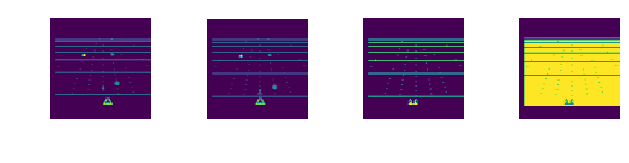}
        
    }
     \subfigure[Beam Rider reward model attention on minimum predicted reward]{
        \includegraphics[width=0.9\linewidth]{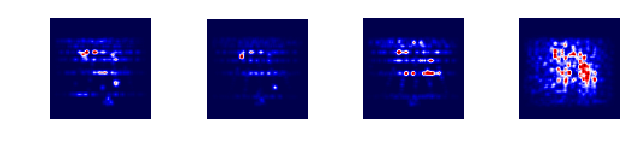}
        
    }
    \caption{Maximum and minimum predicted observations and corresponding attention maps for Beam Rider. The observation with the maximum predicted reward shows successfully destroying an enemy ship, with the network paying attention to the oncoming enemy ships and the shot that was fired to destroy the enemy ship. The observation with minimum predicted reward shows an enemy shot that destroys the player's ship and causes the player to lose a life. The network attends most strongly to the enemy ships but also to the incoming shot.}
    \label{fig:beamrider_minmax}
\end{figure*}

\begin{figure*}
    \centering
    \subfigure[Breakout observation with maximum predicted reward]{
        \includegraphics[width=0.9\linewidth]{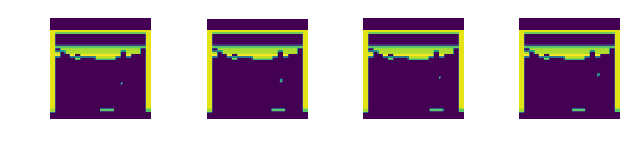}
        \label{subfig:breakout_max}
    }
    \subfigure[Breakout reward model attention on maximum predicted reward]{
        \includegraphics[width=0.9\linewidth]{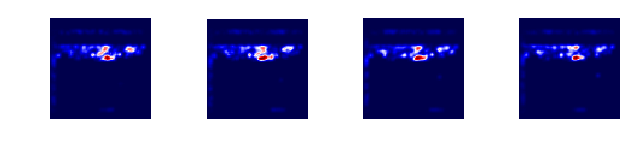}
        \label{subfig:breakout_max_attention}
    }
    \subfigure[Breakout observation with minimum predicted reward]{
        \includegraphics[width=0.9\linewidth]{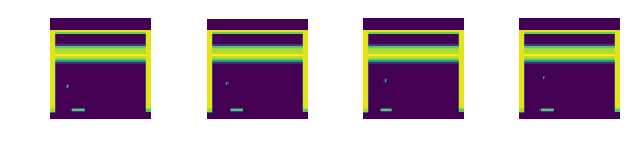}
        \label{subfig:breakout_min}
    }
     \subfigure[Breakout reward model attention on minimum predicted reward]{
        \includegraphics[width=0.9\linewidth]{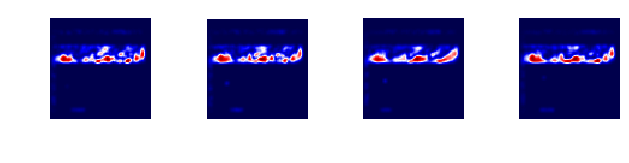}
        \label{subfig:breakout_min_attention}
    }
    \caption{Maximum and minimum predicted observations and corresponding attention maps for Breakout. The observation with maximum predicted reward shows many of the bricks destroyed with the ball on its way to hit another brick. The network has learned to put most of the reward weight on the remaining bricks with some attention on the ball and paddle. The observation with minimum predicted reward is an observation where none of the bricks have been destroyed. The network attention is focused on the bottom layers of bricks. }
    \label{fig:breakout_minmax}
\end{figure*}

\begin{figure*}
    \centering
    \subfigure[Enduro observation with maximum predicted reward]{
        \includegraphics[width=0.9\linewidth]{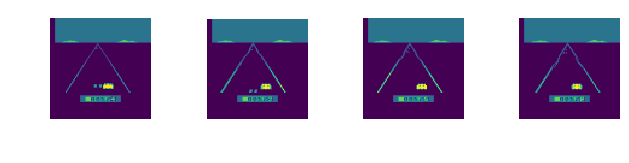}
        
    }
     \subfigure[Enduro reward model attention on maximum predicted reward]{
        \includegraphics[width=0.9\linewidth]{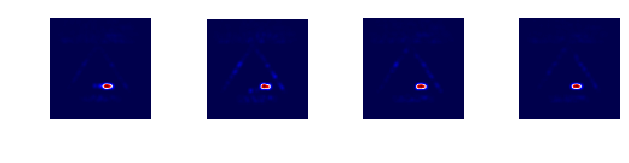}
        
    }
    \subfigure[Enduro observation with minimum predicted reward]{
        \includegraphics[width=0.9\linewidth]{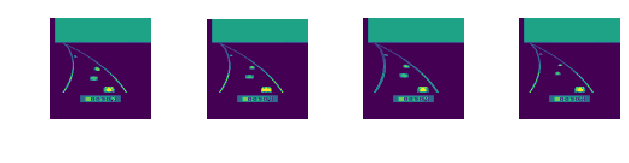}
        
    }
     \subfigure[Enduro reward model attention on minimum predicted reward]{
        \includegraphics[width=0.9\linewidth]{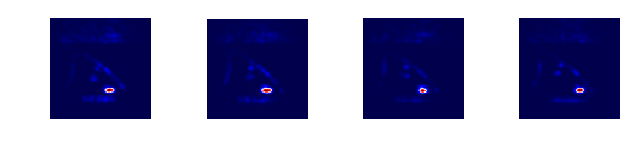}
        
    }
    \caption{Maximum and minimum predicted observations and corresponding attention maps for Enduro. The observation with maximum predicted reward shows the car passing to the right of another car. The network has learned to put attention on the controlled car as well as the sides of the road with some attention on the car being passed and on the odometer. The observation with minimum predicted reward shows the controlled car falling behind other racers, with attention on the other cars, the odometer, and the controlled car.}
    \label{fig:enduro_minmax}
\end{figure*}

\begin{figure*}
    \centering
    \subfigure[Hero observation with maximum predicted reward]{
        \includegraphics[width=0.9\linewidth]{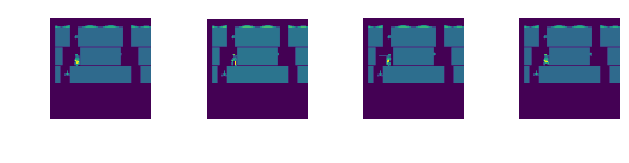}
        
    }
     \subfigure[Hero reward model attention on maximum predicted reward]{
        \includegraphics[width=0.9\linewidth]{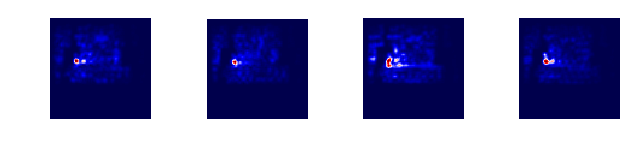}
       
    }
    \subfigure[Hero observation with minimum predicted reward]{
        \includegraphics[width=0.9\linewidth]{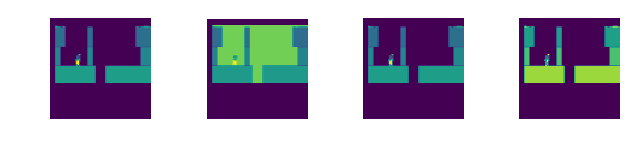}
        
    }
     \subfigure[Hero reward model attention on minimum predicted reward]{
        \includegraphics[width=0.9\linewidth]{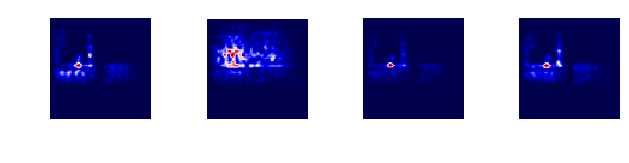}
        
    }
    \caption{Maximum and minimum predicted observations and corresponding attention maps for Hero. The observation with maximum predicted reward is difficult to interpret, but shows the network attending to the controllable character and the shape of the surrounding maze. The observation with minimum predicted reward shows the agent setting off a bomb that kills the main character rather than the wall. The learned reward function attends to the controllable character, the explosion and the wall that was not destroyed.}
    \label{fig:hero_minmax}
\end{figure*}

\begin{figure*}
    \centering
    \subfigure[Pong observation with maximum predicted reward]{
        \includegraphics[width=0.9\linewidth]{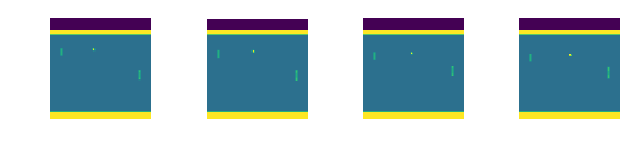}
        
    }
     \subfigure[Pong reward model attention on maximum predicted reward]{
        \includegraphics[width=0.9\linewidth]{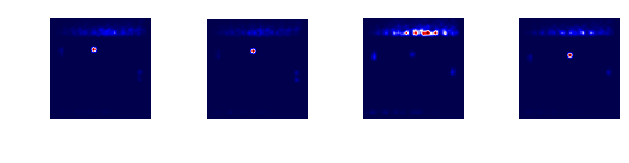}
       
    }
    \subfigure[Pong observation with minimum predicted reward]{
        \includegraphics[width=0.9\linewidth]{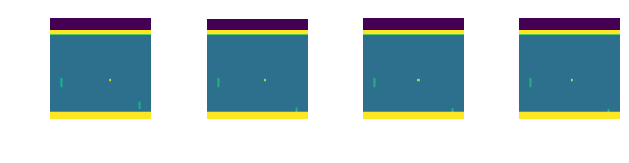}
        
    }
     \subfigure[Pong reward model attention on minimum predicted reward]{
        \includegraphics[width=0.9\linewidth]{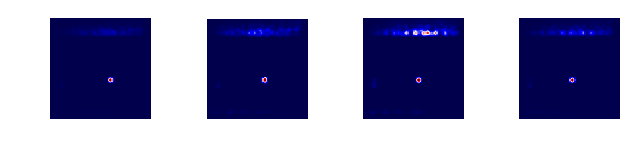}
        
    }
    \caption{Maximum and minimum predicted observations and corresponding attention maps for Pong. The network mainly attends to the ball, with some attention on the paddles.}
    \label{fig:pong_minmax}
\end{figure*}

\begin{figure*}
    \centering
    \subfigure[Q*bert observation with maximum predicted reward]{
        \includegraphics[width=0.9\linewidth]{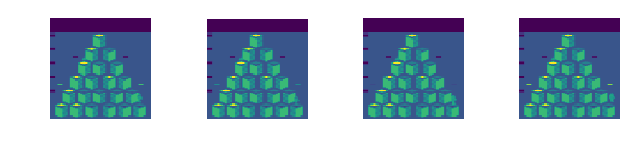}
        
    }
     \subfigure[Q*bert reward model attention on maximum predicted reward]{
        \includegraphics[width=0.9\linewidth]{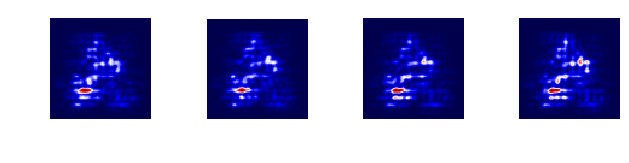}
       
    }
    \subfigure[Q*bert observation with minimum predicted reward]{
        \includegraphics[width=0.9\linewidth]{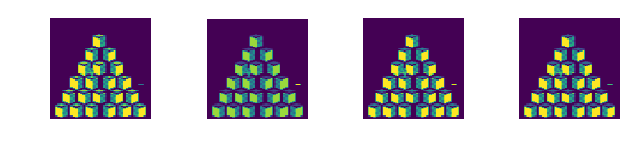}
        
    }
     \subfigure[Q*bert reward model attention on minimum predicted reward]{
        \includegraphics[width=0.9\linewidth]{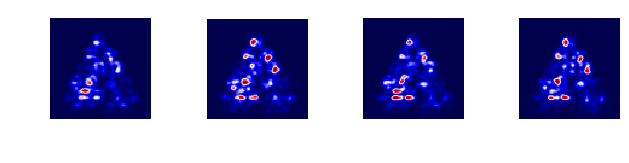}
        
    }
    \caption{Maximum and minimum predicted observations and corresponding attention maps for Q*bert. The observation for the maximum predicted reward shows an observation from the second level of the game where stairs change color from yellow to blue. The observation for the minimum predicted reward is less interpretable. The network attention is focused on the different stairs, but it is difficult to attribute any semantics to the attention maps.}
    \label{fig:qbert_minmax}
\end{figure*}

\begin{figure*}
    \centering
    \subfigure[Seaquest observation with maximum predicted reward]{
        \includegraphics[width=0.9\linewidth]{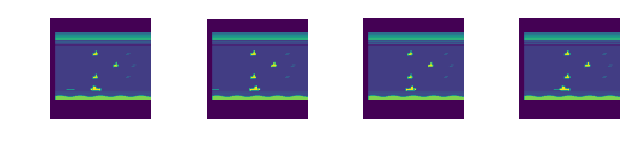}
        
    }
     \subfigure[Seaquest reward model attention on maximum predicted reward]{
        \includegraphics[width=0.9\linewidth]{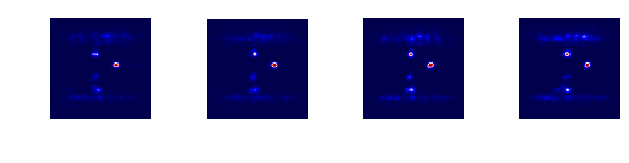}
       
    }
    \subfigure[Seaquest observation with minimum predicted reward]{
        \includegraphics[width=0.9\linewidth]{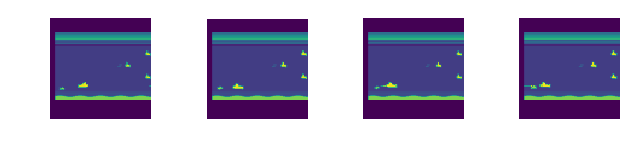}
        
    }
     \subfigure[Seaquest reward model attention on minimum predicted reward]{
        \includegraphics[width=0.9\linewidth]{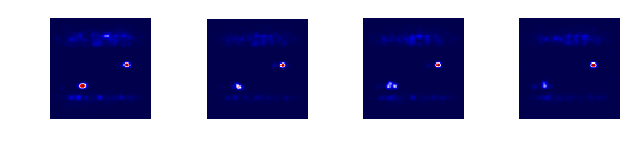}
        
    }
    \caption{Maximum and minimum predicted observations and corresponding attention maps for Seaquest. The observation with maximum predicted reward shows the submarine in a relatively safe area with no immediate threats. The observation with minimum predicted reward shows an enemy that is about to hit the submarine---the submarine fires a shot, but misses. The attention maps show that the network focuses on the nearby enemies and also on the controlled submarine.}
    \label{fig:seaquest_minmax}
\end{figure*}

\begin{figure*}
    \centering
    \captionsetup[subfigure]{labelformat=empty}
    \subfigure[Space Invaders observation with maximum predicted reward]{
        \includegraphics[width=0.9\linewidth]{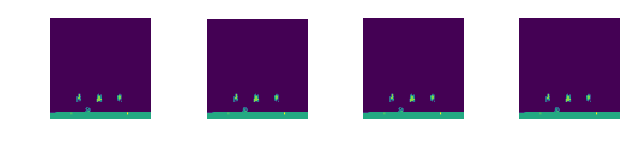}
        
    }
     \subfigure[Space Invaders reward model attention on maximum predicted reward]{
        \includegraphics[width=0.9\linewidth]{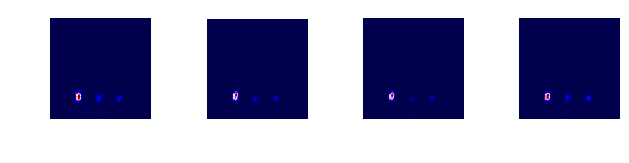}
       
    }
    \subfigure[Space Invaders observation with minimum predicted reward]{
        \includegraphics[width=0.9\linewidth]{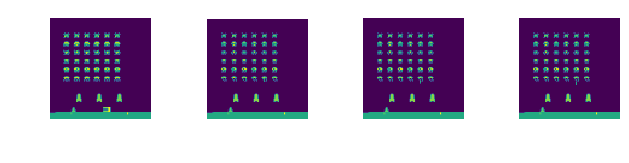}
        
    }
     \subfigure[Space Invaders reward model attention on minimum predicted reward]{
        \includegraphics[width=0.9\linewidth]{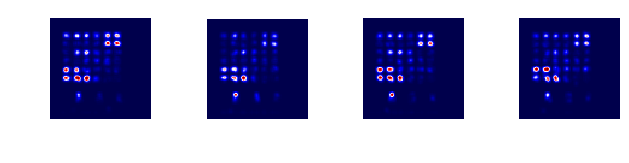}
        
    }
    \caption{Maximum and minimum predicted observations and corresponding attention maps for Space Invaders. The observation with maximum predicted reward shows an observation where all the aliens have been successfully destroyed and the protective barriers are still intact. Note that the agent never observed a demonstration that successfully destroyed all the aliens. The attention map shows that the learned reward function is focused on the barriers, but does not attend to the location of the controlled ship. The observation with minimum predicted reward shows the very start of a game with all aliens still alive. The network attends to the aliens and barriers, with higher weight on the aliens and barrier closest to the space ship.}
    \label{fig:spaceinvaders_minmax}
\end{figure*}

 We generated attention maps for the learned rewards for the Atari domains. We use the method proposed by \citet{greydanus2018visualizing}, which takes a stack of 4 frames and passes a 3x3 mask over each of the frames with a stride of 1. The mask is set to be the default background color for each game. For each masked 3x3 region, we compute the absolute difference in predicted reward when the 3x3 region is not masked and when it is masked. This allows us to measure the influence of different regions of the image on the predicted reward. The sum total of absolute changes in reward for each pixel is used to generate an attention heatmap.  We used the trajectories shown in the extrapolation plots in Figure~4 of the main text and performed a search using the learned reward function to find the observations with minimum and maximum predicted reward. We show the minimum and maximum observations (stacks of four frames) along with the attention heatmaps across all four stacked frames for the learned reward functions in figures \ref{fig:beamrider_minmax}--\ref{fig:spaceinvaders_minmax}. The reward function visualizations suggest that our networks are learning relevant features of the reward function.

 \clearpage

 \begin{table*}
\caption{Best demonstrations and average performance of learned policies for T-REX (ours) and DQfD with active preference learning (DQfD+A) (see \citet{ibarz2018reward} Appendix A.2 and G). Results for T-REX are the best performance over 3 random seeds averaged over 30 trials. Results that exceed the best demonstration are marked with an asterisk (*). Note that T-REX requires at most only 66 pair-wise preference labels ($n(n-1)/2$ for $n=12$ demonstrations), whereas DQfD+A uses between 4--7 demonstrations along with 3.4K labels queried during policy learning. DQfD+A requires action labels on the demonstrations, whereas T-REX learns from observation. }
\label{tab:democomparison}
\vskip 0.15in
\begin{center}
\begin{small}
\begin{tabular}{c|cc|cc}
\toprule
 & \multicolumn{2}{c|}{Best Demonstration Received} & \multicolumn{2}{c}{Average Algorithm Performance} \\
 \midrule
Game &	DQfD+A	& T-REX 	 &
DQfD+A	 & T-REX\\
\midrule
Beam Rider &	19,844 &	1,188 &	4,100 &	*3,335.7 \\
Breakout &	79 &	33 &	*85 &	*221.3\\
Enduro &	803 &	84 &	*1200 &	*586.8  \\
Hero &	99,320 &	13,235 &	35,000 &	0.0 \\
Montezuma's Revenge & 34,900 & - & 3,000 & - \\
Pong &	0 &	-6 &	*19 &	*-2.0 \\
Private Eye & 74,456 & - & 52,000 & - \\
Q*bert &	99,450 &	800 &	14,000 &	*32,345.8 \\
Seaquest &	101,120 &	600 &	500 &	*747.3 \\
Space invaders &	-	 &600	 & -  & 	*1,032.5\\
\bottomrule
\end{tabular}
\end{small}
\end{center}
\vskip -0.1in
\end{table*}

\begin{table*}[t]
\caption{T-REX performance with real novice human demonstrations collected from the Atari Grand Challenge Dataset \cite{kurin2017atari}. Results are the best average performance over 3 random seeds with 30 trials per seed.}
\label{tab:agc_trex}
\vskip 0.15in
\begin{center}
\begin{small}
\begin{tabular}{cccc}
\toprule
& \multicolumn{2}{c}{Novice Human} &\\
Game &  Best & Average & T-REX \\ 
\midrule 
Montezuma's Revenge & \textbf{2,600} &  1,275.0& 0.0\\
Ms Pacman & \textbf{1,360}	 & 818.3	& 550.7\\
Q*bert & 875 & 439.6 & \textbf{6,869.2}\\
Space Invaders & 470 & 290.0 & \textbf{1,092.0}	\\
Video Pinball & 4,210 & 2,864.3 & \textbf{20,000.2}\\
\bottomrule
\end{tabular}
\end{small}
\end{center}
\vskip -0.1in
\end{table*}

\begin{table*}[h]
\caption{Evaluation of T-REX on human rankings collected using Amazon Mechanical Turk. Results are the best average performance over 3 random seeds with 30 trials per seed.}
\label{tab:atariLearningFromHumanRankings}
\vskip 0.15in
\begin{center}
\begin{small}
\begin{tabular}{c|cccc|c}
\toprule
 & \multicolumn{4}{c|}{Human-Ranked Demonstrations} & \\
Game &  Best & Average & Ranking Accuracy & Num. Labels & T-REX avg. perf.\\ 
\midrule 
Beam Rider &	1,332 &	686.0 &	63.0\%& 54 & \textbf{3,457.2}\\
Breakout &	32 &	14.5 & 88.1\%	& 59 & \textbf{253.2}\\
Enduro &	\textbf{84} & 39.8 & 58.6\% & 58 & 0.03\\
Hero &	\textbf{13,235} &	6742 & 77.6\%	& 58 & 2.5\\ 
Pong &	\textbf{-6} &	-15.6 &	79.6\% & 54 & -13.0\\ 
Q*bert &	800 &	627 & 75.9\%	& 58 & \textbf{66,082}\\ 
Seaquest &	600 &	373.3 &	80.4\% & 56 & \textbf{655.3}\\ 
Space Invaders &	600 &	332.9 &	84.7\% & 59 & \textbf{1,005.3}\\ 
\bottomrule
\end{tabular}
\end{small}
\end{center}
\vskip -0.1in
\end{table*}

\clearpage
 
\end{document}